\documentclass[10pt,twocolumn,letterpaper]{article}

\usepackage{cvpr}
\usepackage{times}
\usepackage{epsfig}
\usepackage{graphicx}
\usepackage{amsmath}
\usepackage{amssymb}

\usepackage{color}
\usepackage{multirow}
\usepackage{adjustbox}
\usepackage{subfigure}
\usepackage{paralist} % compactitem, compactenum
\usepackage{enumitem} % adjust enum margin
\usepackage{booktabs} % Publication quality tables
\usepackage{array}

\graphicspath{{figures/}}

\usepackage[pagebackref=true,breaklinks=true,letterpaper=true,colorlinks,citecolor=blue,linkcolor=blue,bookmarks=false]{hyperref}

\newcommand{\Paragraph}[1]{\noindent\paragraph{#1}} % avoid indent for paragraph title

\long\def\ignorethis#1{}

\definecolor{gray}{rgb}{0.35,0.35,0.35}
\definecolor{MyBlue}{rgb}{0,0.2,0.8}
\definecolor{MyRed}{rgb}{0.8,0.2,0}
\definecolor{MyGreen}{rgb}{0.0,0.5,0.1}
\definecolor{MyGray}{rgb}{0.4,0.4,0.4}

\newlength\paramargin
\newlength\figmargin
\newlength\secmargin

\newcolumntype{L}[1]{>{\raggedright\let\newline\\\arraybackslash\hspace{0pt}}m{#1}}
\newcolumntype{C}[1]{>{\centering\let\newline\\\arraybackslash\hspace{0pt}}m{#1}}
\newcolumntype{R}[1]{>{\raggedleft\let\newline\\\arraybackslash\hspace{0pt}}m{#1}}

\newcommand{\figref}[1]{Figure~\ref{fig:#1}}
\newcommand{\tabref}[1]{Table~\ref{tab:#1}}
\newcommand{\eqnref}[1]{\eqref{eq:#1}}

\cvprfinalcopy % *** Uncomment this line for the final submission

\def\cvprPaperID{1531} % *** Enter the CVPR Paper ID here

\ifcvprfinal\fi
\begin{document}

%%%%%%%%% TITLE
\title{Deep Semantic Face Deblurring}

\author{
	Ziyi Shen$^{1}$
	\hspace{25pt}
	Wei-Sheng Lai$^{2}$
	\hspace{25pt}
	Tingfa Xu$^{1}$\thanks{Corresponding author}
	\hspace{25pt}
	Jan Kautz$^{3}$
	\hspace{25pt}
	Ming-Hsuan Yang$^{2,4}$
	\\
	$^1$Beijing Institute of Technology
	\hspace{20pt}
	$^2$University of California, Merced
	%\hspace{20pt}
	\\
	$^3$Nvidia
	\hspace{20pt}
	$^4$Google Cloud
	\\
	{\small
		\url{https://sites.google.com/site/ziyishenmi/cvpr18_face_deblur}
	}
}

\maketitle
%\thispagestyle{empty}

%%%%%%%%%%%%%%%%%%%%%%%%%%%%%%%%%%%%%%%%%%%%%%%%%%%%%%%%%%%%%%%%
%%%%%%%%% Abstract
%%%%%%%%%%%%%%%%%%%%%%%%%%%%%%%%%%%%%%%%%%%%%%%%%%%%%%%%%%%%%%%%
\begin{abstract}

In this paper, we present an effective and efficient face deblurring algorithm by exploiting semantic cues via deep convolutional neural networks (CNNs).
As face images are highly structured and share several key semantic components (e.g., eyes and mouths), the semantic information of a face provides a strong prior for restoration.
As such, we propose to incorporate global semantic priors as input and impose local structure losses to regularize the output within a multi-scale deep CNN.
We train the network with perceptual and adversarial losses to generate photo-realistic results and develop an incremental training strategy to handle random blur kernels in the wild.
Quantitative and qualitative evaluations demonstrate that the proposed face deblurring algorithm restores sharp images with more facial details and performs favorably against state-of-the-art methods in terms of restoration quality, face recognition and execution speed.

\end{abstract}
\vspace{-2mm}

%%%%%%%%%%%%%%%%%%%%%%%%%%%%%%%%%%%%%%%%%%%%%%%%%%%%%%%%%%%%%%%%
%%%%%%%%% Introduction
%%%%%%%%%%%%%%%%%%%%%%%%%%%%%%%%%%%%%%%%%%%%%%%%%%%%%%%%%%%%%%%%
\section{Introduction}
\label{sec:intro}

%% introduction to single image deblurring, natural image priors
Single image deblurring aims to recover a clear image from a single blurred input image.
Conventional methods model the blur process (assuming spatially invariant blur) as the convolution operation between a latent clear image and a blur kernel, and formulate this problem based on the maximum a posteriori (MAP) framework.
As the problem is ill-posed, the state-of-the-art algorithms rely on natural image priors (e.g., $L_0$ gradient~\cite{DBLP:conf/cvpr/XuZJ13} and dark channel prior~\cite{DBLP:conf/cvpr/PanSP016}) to constrain the solution space.

%% domain-specific methods, shortcoming: slow
While existing image priors are effective for deblurring natural images, the underlying assumption may not hold for images from specific categories, e.g., text, face and low-light conditions.
Therefore, numerous approaches exploit domain-specific priors or strategies, such as $L_0$ intensity~\cite{DBLP:journals/pami/PanHS017} for text images and light streaks~\cite{Hu-CVPR-2014} for extremely low-light images.
As face images typically have fewer textures and edges for estimating blur kernels, Pan et al.~\cite{DBLP:conf/eccv/PanHSY14} propose to search a similar face exemplar from an external dataset and extract the contour as reference edges.
However, a similar reference image may not always exist to cover the diversity of face images in the wild.
Furthermore, those methods based on the MAP framework typically entail heavy computational cost due to the iterative optimization of latent images and blur kernels.
The long execution time limits the applications on resource-sensitive platforms, e.g., cloud and mobile devices.

\begin{figure}[t]
	\footnotesize
	\renewcommand{\tabcolsep}{1pt} % adjust horizontal space
	\renewcommand{\arraystretch}{0.5} % adjust vertical space
	\begin{center}
		\begin{tabular}{cccc}
			\includegraphics[width=0.115\textwidth]{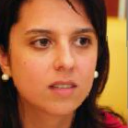} &
			\includegraphics[width=0.115\textwidth]{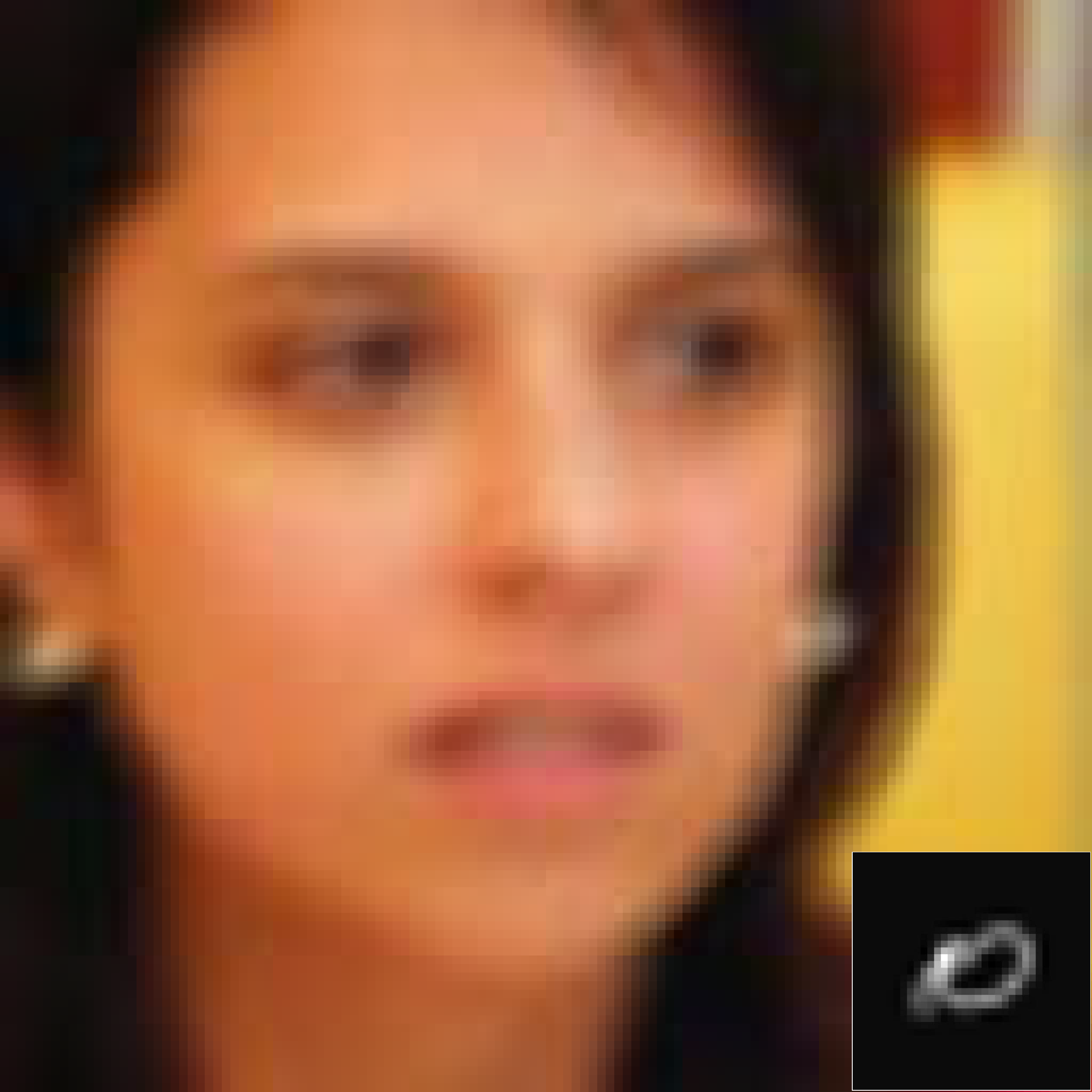} &
			\includegraphics[width=0.115\textwidth]{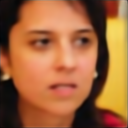} &
			\includegraphics[width=0.115\textwidth]{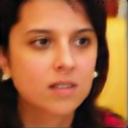}  \\
			\includegraphics[width=0.115\textwidth]{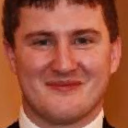} &
			\includegraphics[width=0.115\textwidth]{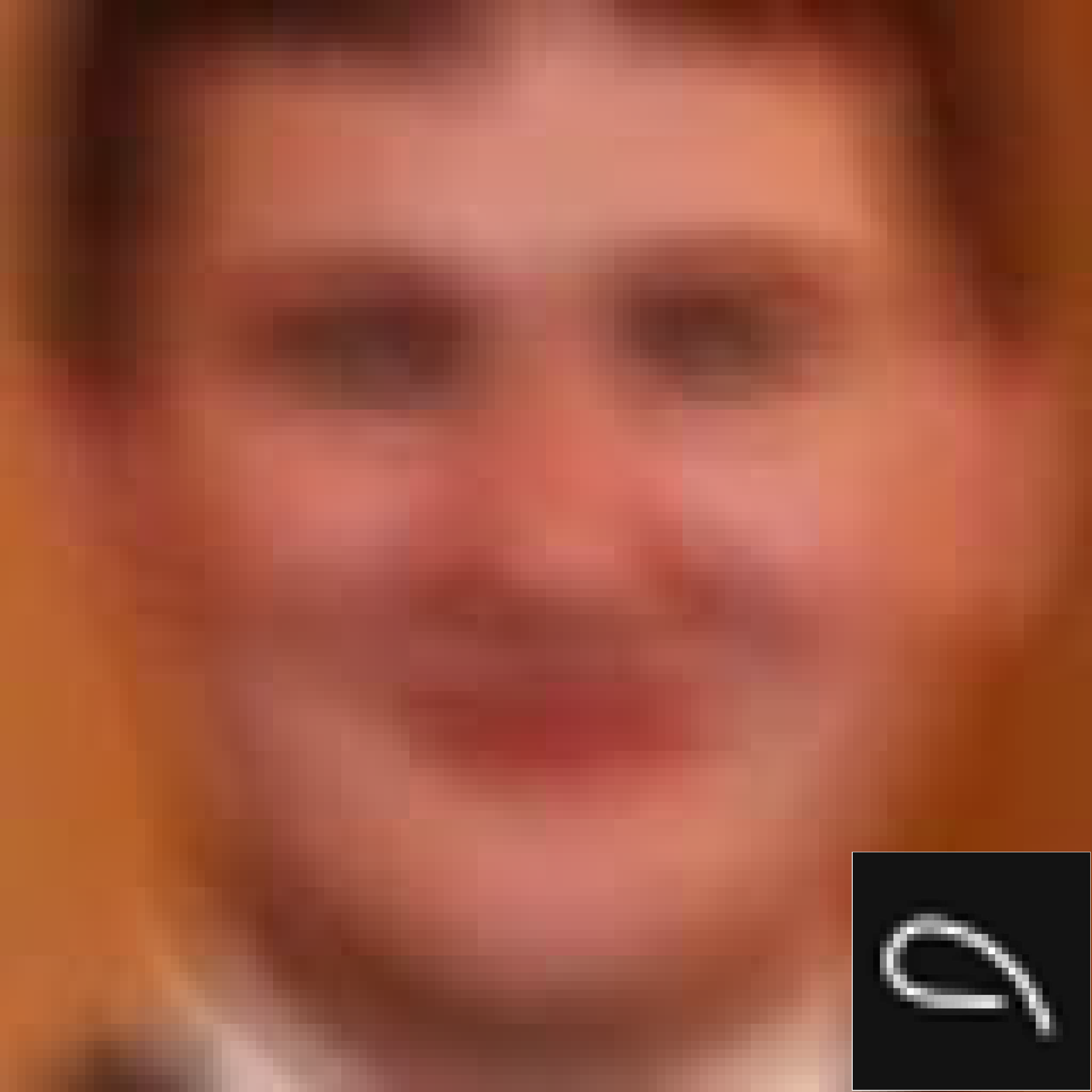} &
			\includegraphics[width=0.115\textwidth]{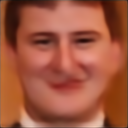} &
			\includegraphics[width=0.115\textwidth]{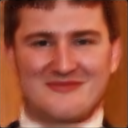} \\
			\includegraphics[width=0.115\textwidth]{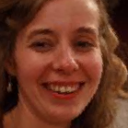} &
			\includegraphics[width=0.115\textwidth]{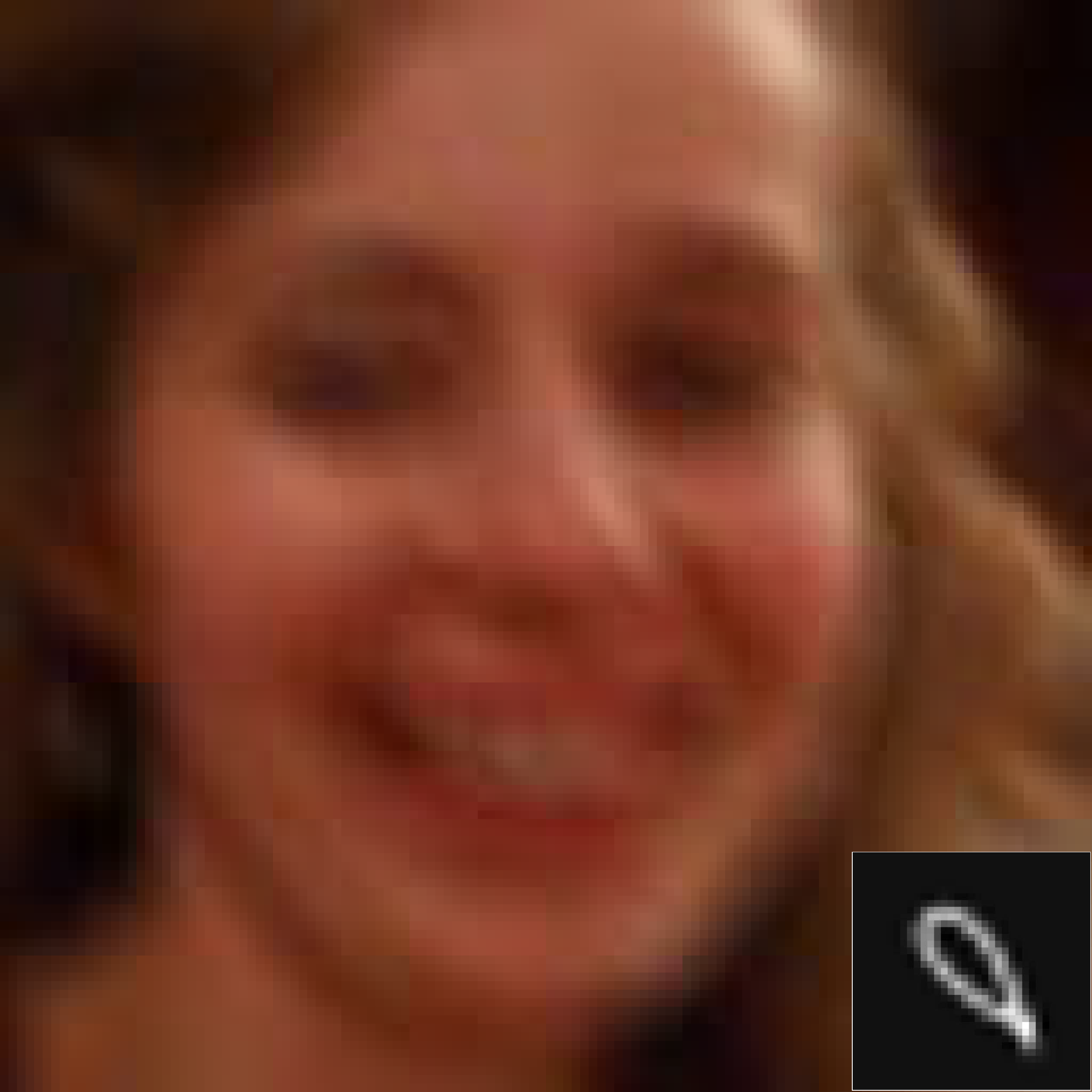} &
			\includegraphics[width=0.115\textwidth]{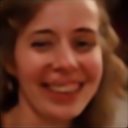} &
			\includegraphics[width=0.115\textwidth]{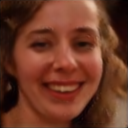} \\
			(a) & (b) & (c) & (d) \\
		\end{tabular}
	\end{center}
	\vspace{-2mm}
	\caption{\textbf{Face deblurring results}.
		We exploit the semantic information of face within an end-to-end deep CNN for face image deblurring.
		(a) Ground truth images
		(b) Blurred images
		(c) Ours w/o semantics
		(d) Ours w/ semantics.
	}
	\label{fig:teaser}
	\vspace{-2mm}
\end{figure}

%% Introduce our goal and CNN-based method
%
In this work, we focus on deblurring face images and propose an efficient as well as effective solution using deep CNNs.
Since face images are highly structured and composed of similar components, the semantic information serves as a strong prior for restoration.
Therefore, we propose to leverage the face semantic labels as global priors and local constraints for deblurring face images.
Specifically, we first generate the semantic labels of blurred input images using a face parsing network.
The face deblurring network then takes the blurred image and semantic labels as input to restore a clear image in a coarse-to-fine manner.
To encourage the network for generating fine details, we further impose a local structure loss on important face components (e.g., eyes, noses, and mouths).
\figref{teaser} shows deblurred examples with and without the proposed semantic priors and losses.
The proposed method is able to reconstruct better facial details than the network trained with only the pixel-wise $L_1$ loss function (i.e., without using semantics).
As our method is end-to-end without any blur kernel estimation or post-processing, the execution time is much shorter than the state-of-the-art MAP-based approaches.

To handle blurred images produced by unknown blur kernels, existing methods typically synthesize blur kernels by modeling the camera trajectories~\cite{DBLP:conf/eccv/Chakrabarti16,DBLP:conf/bmvc/HradisKZS15} and generate a large number of blurred images for training.
Instead of simultaneously using all synthetic blurred images for training, we propose an incremental training strategy by first training the network on a set of small blur kernels and then incorporating larger blur kernels sequentially.
We show that the proposed incremental training strategy facilitates the convergence and improves the performance of our deblurring network
on various sizes of blur kernels.
Finally, we impose a perceptual loss~\cite{Johnson-ECCV-2014} and an adversarial loss~\cite{GAN} to generate photo-realistic deblurred results.

We make the following contributions in this work:
\begin{compactitem}
\item We propose a deep multi-scale CNN that exploits global semantic priors and local structural constraints for face image deblurring.
\item We present an incremental strategy to train CNNs to better handle unknown motion blur kernels.
\item We demonstrate that the proposed method performs favorably against state-of-the-art deblurring approaches in terms of restoration quality, face recognition and execution speed.
\end{compactitem}

%%%%%%%%%%%%%%%%%%%%%%%%%%%%%%%%%%%%%%%%%%%%%%%%%%%%%%%%%%%%%%%%
%%%%%%%%% Related Work
%%%%%%%%%%%%%%%%%%%%%%%%%%%%%%%%%%%%%%%%%%%%%%%%%%%%%%%%%%%%%%%%
\section{Related Work}
\label{sec:related}

Single image deblurring can be categorized into non-blind and blind deblurring based on whether the blur kernel is available or not.
We focus our discussion on blind image deblurring in this section.

\vspace{-4mm}
\Paragraph{Generic methods.}
The recent progress in single image blind deblurring can be attributed to the development of effective natural image priors, including sparse image gradient prior~\cite{DBLP:journals/tog/FergusSHRF06,DBLP:conf/cvpr/LevinWDF09}, normalized sparsity measure~\cite{DBLP:conf/cvpr/KrishnanTF11}, patch prior~\cite{Sun-ICCP-2013}, $L_0$ gradient~\cite{DBLP:conf/cvpr/XuZJ13}, color-line model~\cite{Lai-CVPR-2015}, low-rank prior~\cite{ren2016image}, self-similarity~\cite{Michaeli-ECCV-2014} and dark channel prior~\cite{DBLP:conf/cvpr/PanSP016}.
Through optimizing the image priors within the MAP framework, those approaches implicitly restore strong edges for estimating the blur kernels and latent sharp images.
However, solving complex non-linear priors involve several optimization steps and entail high computational loads.
As such, edge-selection based methods~\cite{DBLP:journals/tog/ChoL09,DBLP:conf/eccv/XuJ10} adopt simple image priors (e.g., $L_2$ gradients) with image filters (e.g., shock filter) to explicitly restore or select strong edges.
While generic image deblurring methods demonstrate state-of-the-art performance, face images have different statistical properties than natural scenes and cannot be restored well using the above approaches.

\vspace{-4mm}
\Paragraph{Domain-specific methods.}
To handle images from specific categories, several domain-specific image deblurring approaches have been developed.
Pan et al.~\cite{DBLP:journals/pami/PanHS017} introduce the $L_0$-regularized priors on both intensity and image gradients for text image deblurring as text images usually contain nearly uniform intensity.
To handle extreme cases such as low-light images, Hu et al.~\cite{Hu-CVPR-2014} detect the light streaks in images for estimating blur kernels.
Anwar et al.~\cite{Anwar-ICCV-2015} propose a frequency-domain class-specific prior to restore the band-pass frequency components.
In addition, a number of approaches use reference images as guidance for non-blind~\cite{Sun-ECCV-2014} and blind deblurring~\cite{Hacohen-ICCV-2013}.
However, the performance of such methods hinges on the similarity of the reference images and quality of dense correspondence.

As face images have fewer textures and edges, existing algorithms based on implicit or explicit edge restoration are less effective.
Pan et al.~\cite{DBLP:conf/eccv/PanHSY14} search for similar faces from a face dataset and extract reference exemplar contours for estimating blur kernels.
However, this approach requires manual annotations of the face contours and involves computationally expensive optimization processes of blur kernels and latent images in the MAP framework.
In contrast, we train an end-to-end deep CNN to bypass the blur kernel estimation step and do not use any reference images or manual annotations when deblurring an image.

\vspace{-4mm}
\Paragraph{CNN-based methods.}
Deep CNNs have been adopted for several image restoration tasks, such as denoising~\cite{Mao-NIPS-2016}, JPEG deblocking~\cite{Dong-ICCV-2015}, dehazing~\cite{ren2016single} and super-resolution~\cite{VDSR,LapSRN}.
Recent approaches apply deep CNNs for image deblurring in several aspects, including non-blind deconvolution~\cite{Schuler-CVPR-2013,DBLP:conf/nips/XuRLJ14,Zhang_2017_CVPR}, blur kernel estimation~\cite{DBLP:journals/pami/SchulerHHS16} and dynamic scene deblurring~\cite{Nah_2017_CVPR}.
Chakrabarti et al.~\cite{DBLP:conf/eccv/Chakrabarti16} train a deep network to predict the Fourier coefficients of a deconvolution filter.
%
%%MH: check this statement.
Despite computational efficiency, these CNN-based methods do not perform as well as the state-of-the-art MAP-based approaches, especially on large motion kernels.

%Domain-specific CNN-based methods include the work of Hradi\v{s} et al.~\cite{DBLP:conf/bmvc/HradisKZS15} and Xu et al.~\cite{Xu-ICCV-2017}.
%
Since text images usually contain uniform intensities with fewer texture regions, an end-to-end deep network~\cite{DBLP:conf/bmvc/HradisKZS15} performs well, especially under large noise levels.
Xu et al.~\cite{Xu-ICCV-2017} aim to jointly deblur and super-resolve low-resolution blurred face and text images, which are typically degraded by Gaussian-like blur kernels.
In this work, we focus on deblurring face images from complex motion blur.
We exploit global and local semantic cues as well as perceptual~\cite{Johnson-ECCV-2014} and adversarial~\cite{GAN} losses to restore photo-realistic face images with fine details.

\begin{figure*}
	\centering
	\footnotesize
	\includegraphics[width=0.8\textwidth]{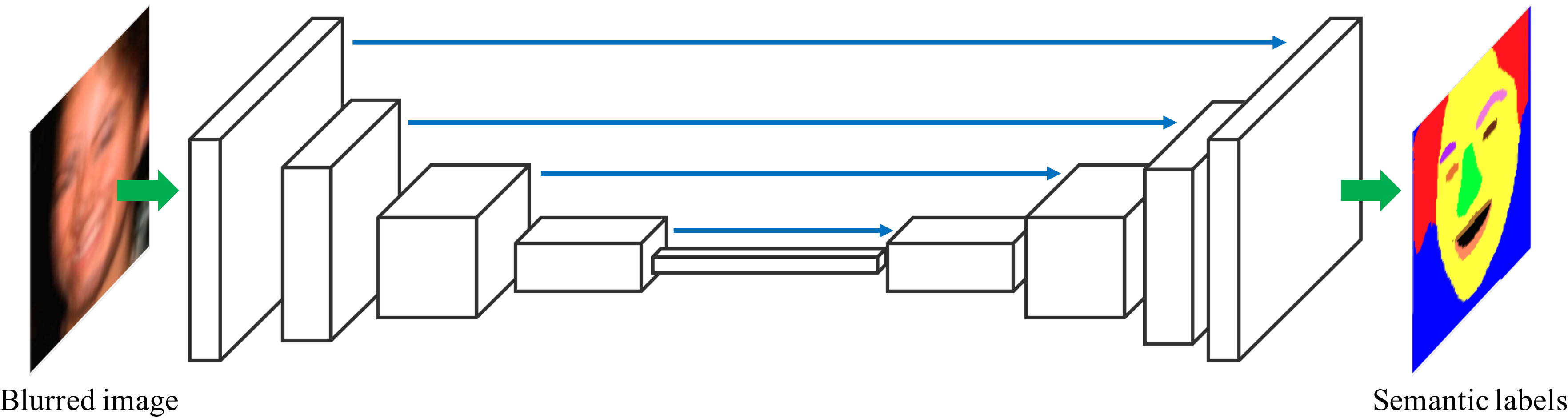}
	\\
	(a) Semantic face parsing network
	\\
	\includegraphics[width=1.0\textwidth]{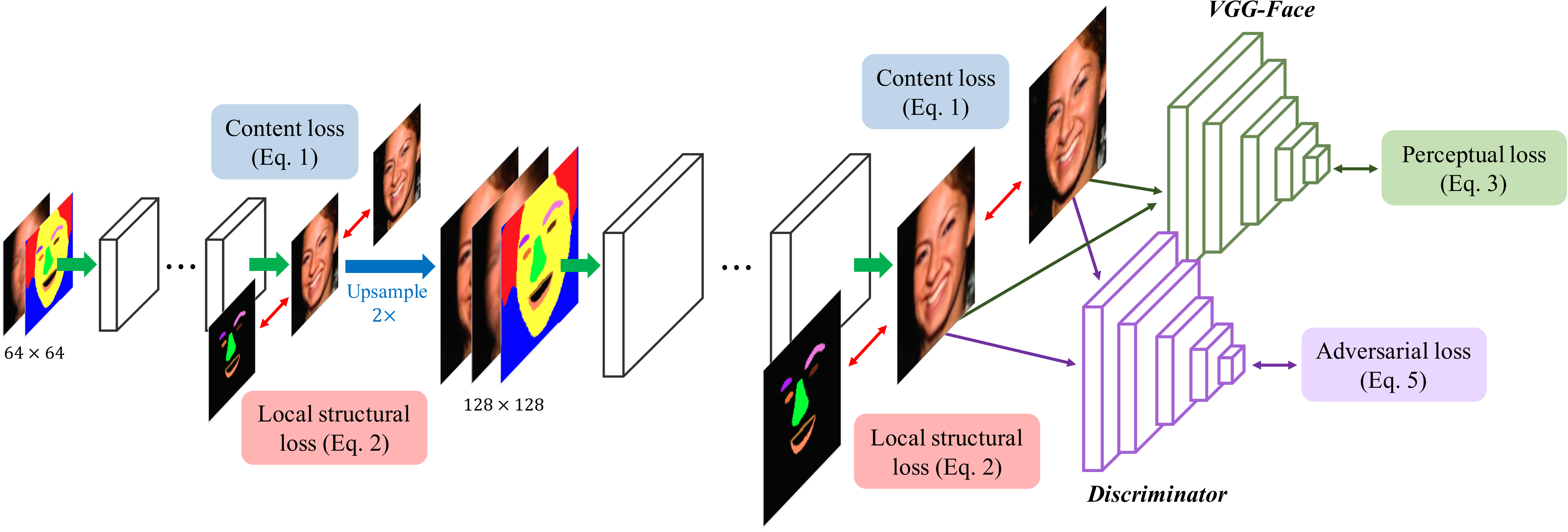}
	\\
	(b) Face deblurring network
	\vspace{1mm}
	\caption{
		\textbf{Overview of the proposed semantic face deblurring network.}
		The proposed network consists of two sub-networks: a semantic face parsing network and a multi-scale deblurring network.
		The face parsing network generates the semantic labels of the input blurred images.
		The multi-scale deblurring network has two scales.
		We concatenate the blurred image and semantic labels as the input to the first scale.
		At the second scale, the input is the concatenation of the upsampled deblurred image from the first scale, the blurred image and the corresponding semantic labels.
		Each scale of the deblurring network receives the supervision from the pixel-wise content loss and local structural losses.
		We impose the perceptual and adversarial losses at the output of the second scale.
		}
	\label{fig:network}
\end{figure*}

%%%%%%%%%%%%%%%%%%%%%%%%%%%%%%%%%%%%%%%%%%%%%%%%%%%%%%%%%%%%%%%%
%%%%%%%%% Method
%%%%%%%%%%%%%%%%%%%%%%%%%%%%%%%%%%%%%%%%%%%%%%%%%%%%%%%%%%%%%%%%
\section{Semantic Face Deblurring}
\label{sec:algorithm}
In this section, we describe the design methodology of the proposed semantic face deblurring approach.
We exploit the semantic labels from a face parsing network as the global semantic priors and local structural losses within a deep multi-scale CNN.
We then train the proposed network jointly with perceptual and adversarial losses to generate photo-realistic deblurred results.

\subsection{Face deblurring network}
\label{sec:deblur_network}
%
%We first describe the architecture of our deblurring network.
%
We use a multi-scale network similar to that from Nah et al.~\cite{Nah_2017_CVPR}, but with several differences.
First, as face images typically have a spatial resolution of $128 \times 128$ or less, we use only 2 scales instead of 3 scales for natural images in~\cite{Nah_2017_CVPR}.
Second, we use fewer ResBlocks (reduce from 19 to 6) and larger filter size ($11 \times 11$) at the first convolutional layer to increase the receptive field.
Finally, we introduce additional inputs from semantic face parsing as global priors and impose local structural constraints on the output at each scale.

\subsection{Global semantic priors}
\label{sec:global_semantic}
We propose to utilize the semantic parsing information as a global prior for face deblurring.
Given a blurred image, we first use a face parsing network ~\cite{DBLP:conf/cvpr/LiuYHY15} to extract the semantic labels.
We then concatenate the probability maps of the semantic labels with the blurred face image as the input to our deblurring network.
The input to the first scale of the deblurring network has a spatial resolution of $64 \times 64$ and a total of 14 channels (3-channel RGB image and 11-channel semantic probabilities).
The deblurred image of the first scale is then upsampled by $2\times$ through a transposed convolutional layer.
The input of the second scale has a spatial resolution of $128 \times 128$ and a total of 17 channels, including the upsampled deblurred image, the blurred image, and the corresponding semantic probabilities.
\figref{network} shows an overview of our face parsing and deblurring network.
The semantic labels encode the essential appearance information and rough locations of the facial components (e.g., eyes, noses and mouths) and serve as a strong global prior for reconstructing the deblurred face image.

\subsection{Local structural constraints}
\label{sec:local_structure}
We use the pixel-wise $L_1$ robust function as the content loss of our face deblurring network:
\begin{equation}
	\mathcal{L}_c = \left\| \mathcal{G}(B, \mathcal{P}(B)) - I \right\|_1,
	\label{eq:L1_loss}
\end{equation}
where $\mathcal{P}$ and $\mathcal{G}$ denote the face parsing and deblurring networks.
In addition, $B$ and $I$ are the blurred and ground truth clear images, respectively.
However, the key components (e.g., eyes, lips and mouths) on faces are typically small and cannot be well reconstructed by solely minimizing the content loss on the whole face image.
As human vision is more sensitive to the artifacts on key components, we propose to impose local structural losses:
\begin{equation}
	\mathcal{L}_s = \sum_{k = 1}^K \left\| \mathbb{M}_k(\mathcal{P}(B)) \odot \left( \mathcal{G}(B, \mathcal{P}(B)) - I \right)  \right\|_1,
	\label{eq:structural_loss}
\end{equation}
where $\mathbb{M}_k$ denotes the structural mask of the $k$-th component and $\odot$ is the element-wise multiplication.
%
%%MH: tooth, too? I know other components from parsing results
We apply the local structural losses on eyebrows, eyes, noses, lips and teeth.
The local structural losses enforce the deblurring network to restore more details on those key components.

\subsection{Generating photo-realistic face images}
\label{sec:photorealistic}

As pixel-wise $L_2$ or $L_1$ loss functions typically lead to overly-smooth results, we introduce a perceptual loss~\cite{Johnson-ECCV-2014} and an adversarial loss~\cite{GAN} to optimize our deblurring network and generate photo-realistic deblurred results.

\Paragraph{Perceptual loss.}
The perceptual loss has been adopted in style transfer~\cite{DBLP:conf/nips/GatysEB15,Johnson-ECCV-2014}, image super-resolution~\cite{Ledig_2017_CVPR} and image synthesis~\cite{Chen-ICCV-2017}.
The perceptual loss aims to measure the similarity in the high dimensional feature space of a pre-trained loss network (e.g., VGG16~\cite{conf/ICLR/Simonyan15}).
Given the input image $x$, we denote $\phi_l(x)$ as the activation at the $l$-th layer of the loss network $\phi$.
The perceptual loss is then defined as:
\begin{equation}
	\mathcal{L}_p = \sum_{l} \left\| \phi_l(\mathcal{G}(B)) - \phi_l(I) \right\|_2^2.
	\label{eq:perceptual_loss}
\end{equation}
We compute the perceptual loss on the Pool2 and Pool5 layers of the pre-trained VGG-Face~\cite{VGG-Face} network.

\Paragraph{Adversarial loss.}
The adversarial training framework has been shown effective to synthesize realistic images~\cite{GAN,Ledig_2017_CVPR,Nah_2017_CVPR}.
We treat our face deblurring network as the generator and construct a discriminator based on the architecture of DCGAN~\cite{DCGAN}.
The goal of the discriminator $\mathcal{D}$ is to distinguish the real image from the output of the generator.
The generator $\mathcal{G}$ aims to generate images as real as possible to fool the discriminator.
The adversarial training is formulated as solving the following min-max problem:
\begin{align}
	\underset{\mathcal{G}}{\mathop{\min }}\,
	\underset{\mathcal{D}}{\mathop{\max }}\,
	\mathbb{E} \left[ \log \mathcal{D}(I) \right]
	+ \mathbb{E} \left[ \log (1-\mathcal{D}( \mathcal{G}(B) )) \right].
	\label{eq:adversarial_loss}
\end{align}
When updating the generator, the adversarial loss is:
\begin{equation}
	\mathcal{L}_{\text{adv}} = - \log \mathcal{D}( \mathcal{G} (B)).
	\label{eq:adv_G}
\end{equation}
Our discriminator takes an input image with a size of $128 \times 128$ and has 6 strided convolutional layers followed by the ReLU activation function.
In the last layer, we use the sigmoid function to output a single scalar as the probability to be a real image.

\Paragraph{Overall loss function.}
The overall loss function for training our face deblurring network consists of the content loss, local structural losses, perceptual loss and adversarial loss:
\begin{equation}
	\mathcal{L} = \mathcal{L}_c + \lambda_s \mathcal{L}_s + \lambda_p \mathcal{L}_p + \lambda_{\text{adv}} \mathcal{L}_{\text{adv}},
\end{equation}
where $\lambda_s$, $\lambda_p$ and $\lambda_{\text{adv}}$ are the weights to balance the local structural losses, perceptual loss and adversarial loss, respectively.
In this work, we empirically set the weights to $\lambda_s = 50$, $\lambda_p = 1e^{-5}$ and $\lambda_{\text{adv}} = 5e^{-5}$.
We apply the content and local structural losses at all scales of the deblurring network while only adopt the perceptual and adversarial losses at the finest scale (i.e., second scale).

\subsection{Implementation details}

We use a variant of Liu et al.~\cite{DBLP:conf/cvpr/LiuYHY15} as our semantic face parsing network, which is an encoder-decoder architecture with skip connections from the encoder to the decoder (see~\figref{network}(a)).
Our face deblurring network has two scales, where each scale has 6 ResBlocks and a total of 18 convolutional layers.
All convolutional layers except the first layer have the kernel size of $5 \times 5$ and 64 channels.
The upsampling layer uses a $4 \times 4$ transposed convolutional layer to upsample the image by $2\times$.
The detailed architecture of our face deblurring network is described in the supplementary material.

We implement our network using the MatConvNet toolbox~\cite{matconvnet}.
We use a batch size of 16 and set the learning rate to $5e^{-6}$ when training the parsing network and $4e^{-5}$ when training the deblurring network.
The parsing network converges within 60,000 iterations and the training takes less than one day.
We train the deblurring network for 17 million iterations, which takes about 5 days on an NVIDIA Titan X GPU.
We note that we first train the semantic face parsing network until convergence.
We then fix the parsing network while training the deblurring network.
%
%MH: add these statements
%All the source code will be made available to the public.

%%%%%%%%%%%%%%%%%%%%%%%%%%%%%%%%%%%%%%%%%%%%%%%%%%%%%%%%%%%%%%%%
%%%%%%%%% Experiment
%%%%%%%%%%%%%%%%%%%%%%%%%%%%%%%%%%%%%%%%%%%%%%%%%%%%%%%%%%%%%%%%
\section{Experimental Results}
\label{sec:results}

In this section, we first describe the training and test datasets used in our experiments.
We then analyze the performance of the semantic face parsing network and face deblurring network, describe our incremental training strategy to handle random blur kernels, and finally compare with state-of-the-art deblurring algorithms.

\subsection{Datasets}
We use the Helen dataset~\cite{helen}, which has ground truth face semantic labels, for training our semantic face parsing network.
The Helen dataset consists of 2,000 training images and 330 validation images.
We use the method of Sun et al.~\cite{DBLP:conf/cvpr/SunWT13} to detect the facial key points and align all face images using the method of Kae et al.~\cite{DBLP:conf/cvpr/KaeSLL13}.
During training, we apply data augmentation using affine transformations to avoid over-fitting.

To train the deblurring network, we collect training images from the Helen dataset~\cite{helen} (2,000 images), CMU PIE dataset~\cite{PIE} (2,164 images) and CelebA dataset~\cite{CelebA} (2,300 images) as our training data.
We synthesize 20,000 motion blur kernels from random 3D camera trajectories~\cite{DBLP:journals/tip/BoracchiF12}.
The size of blur kernels range from $13 \times 13$ to $27 \times 27$.
By convolving the clear images with blur kernels and adding Gaussian noise with $\sigma = 0.01$, we obtain 130 million blurred images for training.

%\subsection{Test datasets}
%
In addition to the training set, we synthesize another 80 random blur kernels, which are different from the 20,000 blur kernels used for training.
We collect 100 clear face images from the validation set of the Helen and CelebA datasets, respectively.
There are a total of 16,000 blurred images for testing.

\subsection{Semantic face parsing}
\begin{figure}
	\footnotesize
	\renewcommand{\tabcolsep}{1pt} % adjust horizontal space
	\renewcommand{\arraystretch}{0.5} % adjust vertical space
	\begin{center}
		\begin{tabular}{cccc}
			\includegraphics[width=0.115\textwidth]{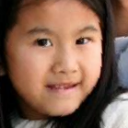} &
			\includegraphics[width=0.115\textwidth]{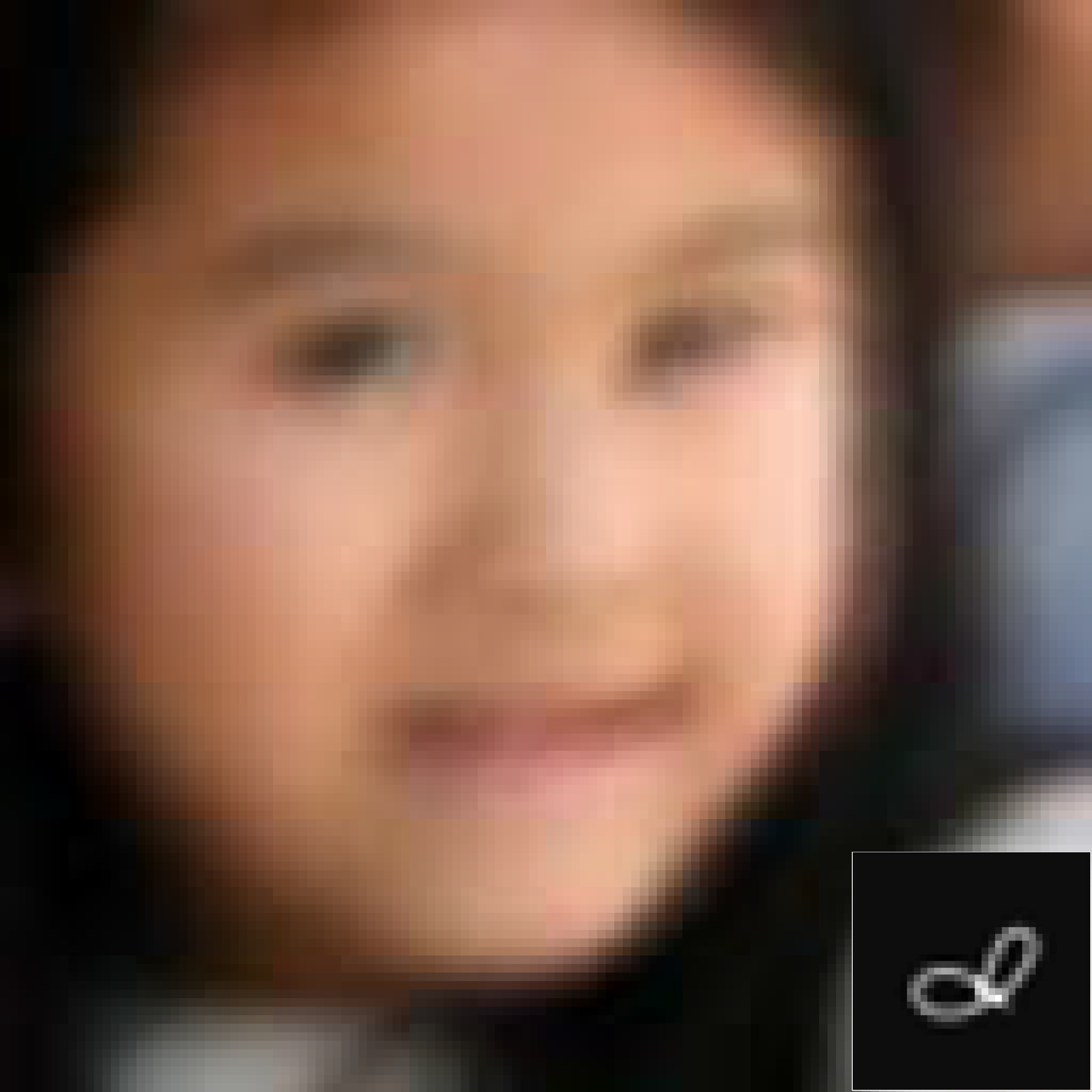} &
			\includegraphics[width=0.115\textwidth]{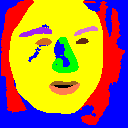} &
            \includegraphics[width=0.115\textwidth]{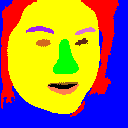} \\
			\includegraphics[width=0.115\textwidth]{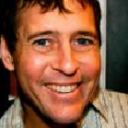} &
			\includegraphics[width=0.115\textwidth]{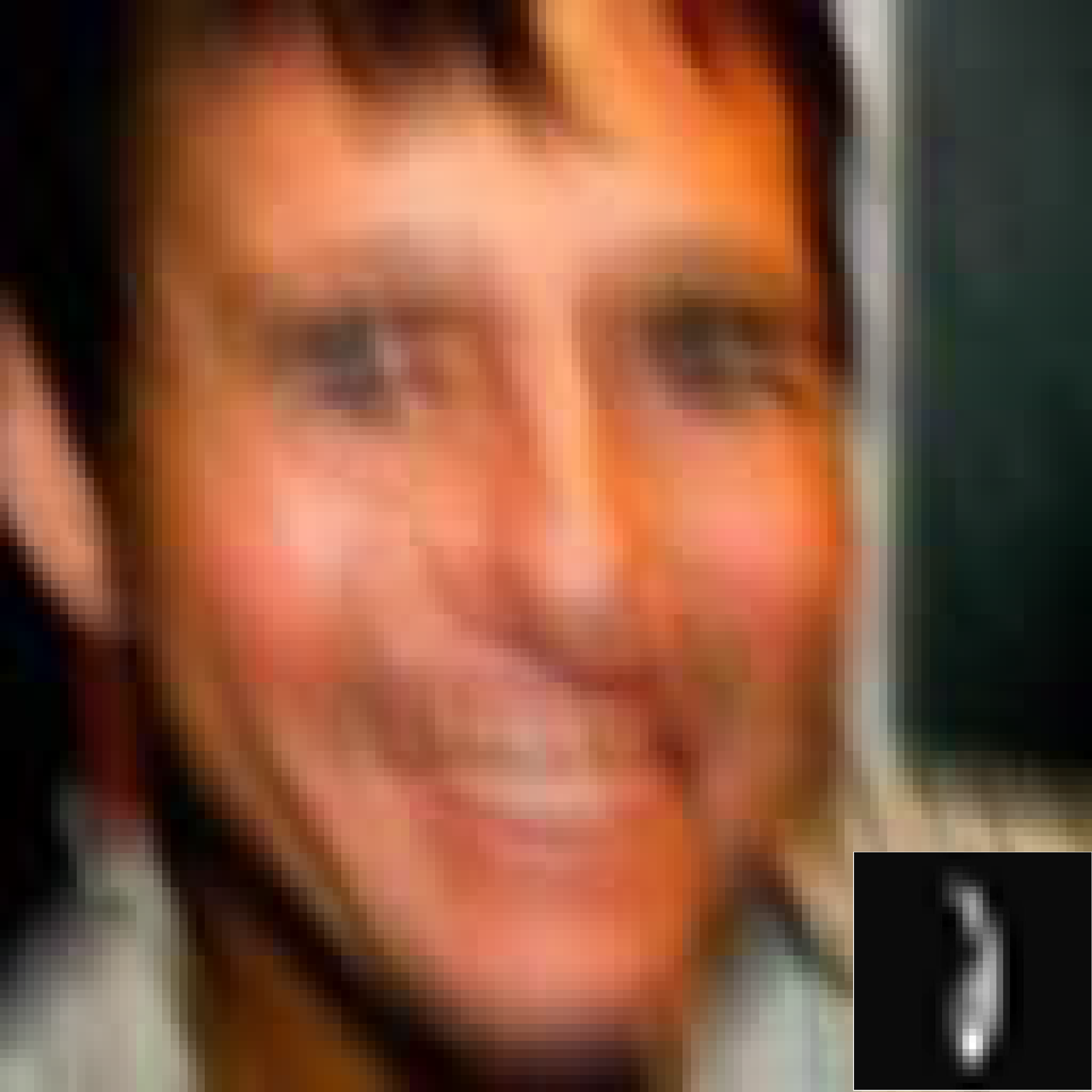} &
            \includegraphics[width=0.115\textwidth]{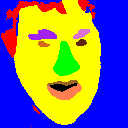} &
			\includegraphics[width=0.115\textwidth]{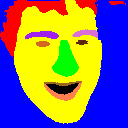} \\
            (a) & (b) & (c) & (d)
		\end{tabular}
	\end{center}
	\vspace{-2mm}
	\caption{
		\textbf{Labeling results of our semantic face parsing network.}
		(a) Ground truth images
		(b) Blurred images
		(c) Results from pre-trained model (trained on clear images)
		(d) Results from fine-tuned model (fine-tuned on blurred images).
	}
	\label{fig:face_parsing}
	\vspace{-4mm}
\end{figure}

We first validate the performance of our semantic face parsing network.
We use the images from the Helen dataset for training and evaluate the F-scores of each facial component on the Helen validation set.
We report the performance on clear and blurred images in~\tabref{face_parsing}.
Due to motion blur, the face parsing network does not perform well on blurred images, especially for small and thin components, e.g., eyebrows, lips, and teeth.
We further fine-tune the parsing network on blurred images to improve the performance.
\figref{face_parsing} shows the parsing results before and after fine-tuning on blurred images.
The fine-tuned model is more robust to motion blur and parses facial components well.

\begin{table}
	\centering
	\footnotesize
	\caption{\textbf{Performance of our semantic face parsing network.}
		We measure the F-score on each facial component.
		``Pre-trained'' model denotes the network trained on clear images.
		``Fine-tuned'' model is the network fine-tuned on blurred images.
	}
	\label{tab:face_parsing}
	\vspace{1mm}
	\begin{tabular}{c|ccc}
		\toprule
		Input image & Clear & \multicolumn{2}{c}{Blurred} \\
		Evaluated model & Pre-trained & Pre-trained & Fine-tuned \\
		\midrule
		face            & 0.923 & 0.891 & 0.896 \\
		left eyebrow    & 0.730 & 0.574 & 0.596 \\
		right eyebrow   & 0.731 & 0.581 & 0.618 \\
		left eye        & 0.748 & 0.602 & 0.677 \\
		right eye       & 0.786 & 0.630 & 0.608 \\
		nose            & 0.893 & 0.875 & 0.855 \\
		upper lip       & 0.645 & 0.489 & 0.477 \\
		lower lip       & 0.744 & 0.605 & 0.650 \\
		teeth           & 0.451 & 0.303 & 0.369 \\
		hair            & 0.557 & 0.481 & 0.499 \\
		average         & 0.721 & 0.603 & 0.625 \\
		\bottomrule
	\end{tabular}
	\vspace{-2mm}
\end{table}

\subsection{Face image deblurring}
In this section, we evaluate the effect of using semantic information on face image deblurring, describe our incremental training strategy for handling random blur kernels, and compare with state-of-the-art deblurring methods.

\vspace{-2mm}
\Paragraph{Effect of semantic parsing.}
We train a baseline model using only the content loss function~\eqnref{L1_loss}.
We then train another two models by first introducing the semantic labels as input priors and then including the local structural losses~\eqnref{structural_loss}.

\begin{figure}
	\footnotesize
	\renewcommand{\tabcolsep}{1pt} % adjust horizontal space
	\renewcommand{\arraystretch}{0.5} % adjust vertical space
	\centering
	\begin{tabular}{ccccc}
            \includegraphics[width=0.09\textwidth]{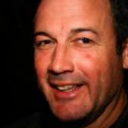} &
			\includegraphics[width=0.09\textwidth]{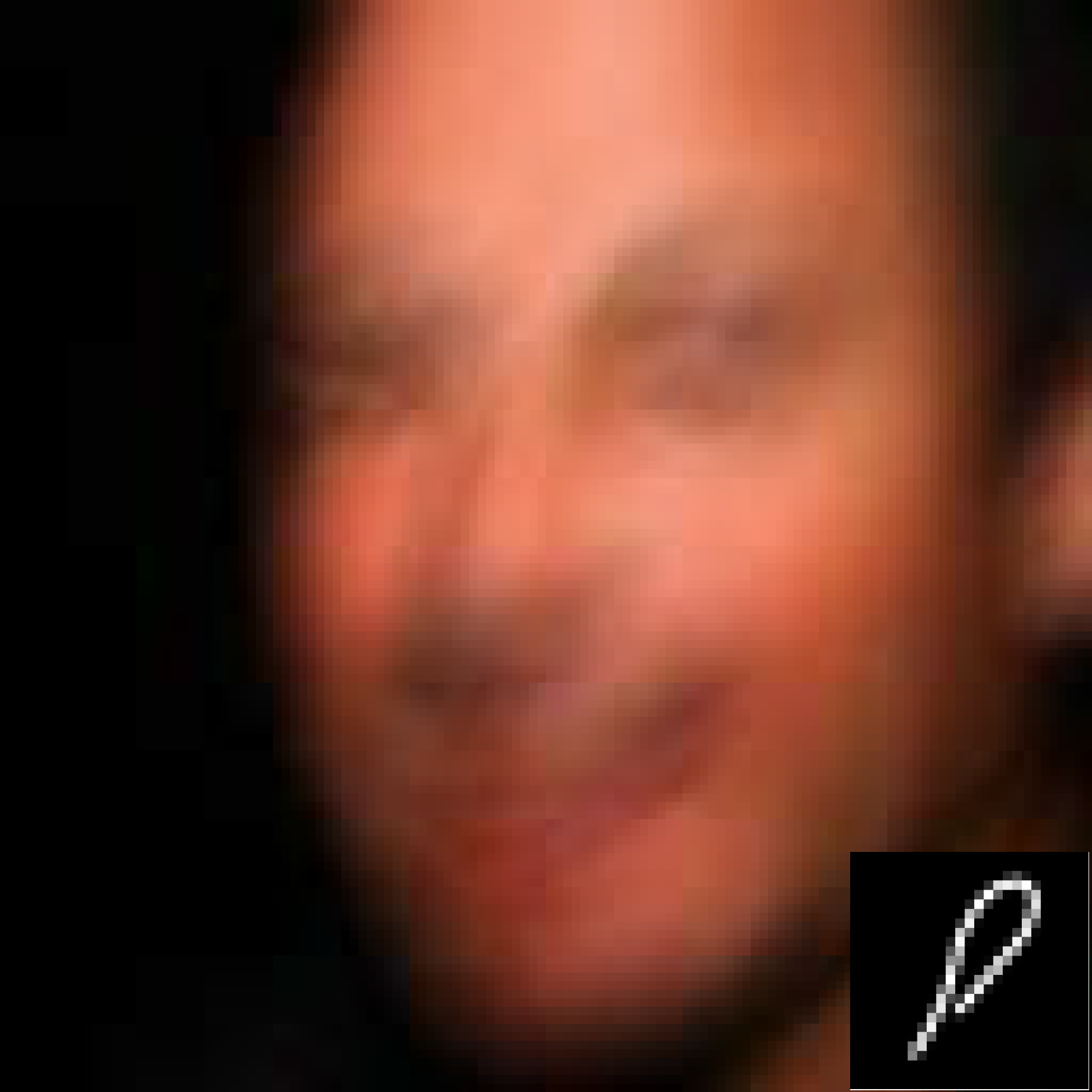} &
            \includegraphics[width=0.09\textwidth]{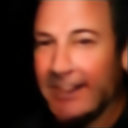} &
			\includegraphics[width=0.09\textwidth]{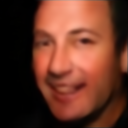} &
			\includegraphics[width=0.09\textwidth]{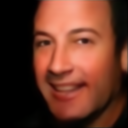}  \\
            \includegraphics[width=0.09\textwidth]{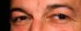} &
			\includegraphics[width=0.09\textwidth]{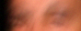} &
			\includegraphics[width=0.09\textwidth]{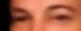} &
            \includegraphics[width=0.09\textwidth]{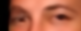} &
			\includegraphics[width=0.09\textwidth]{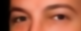}  \\
            \includegraphics[width=0.09\textwidth]{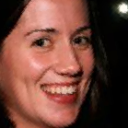} &
			\includegraphics[width=0.09\textwidth]{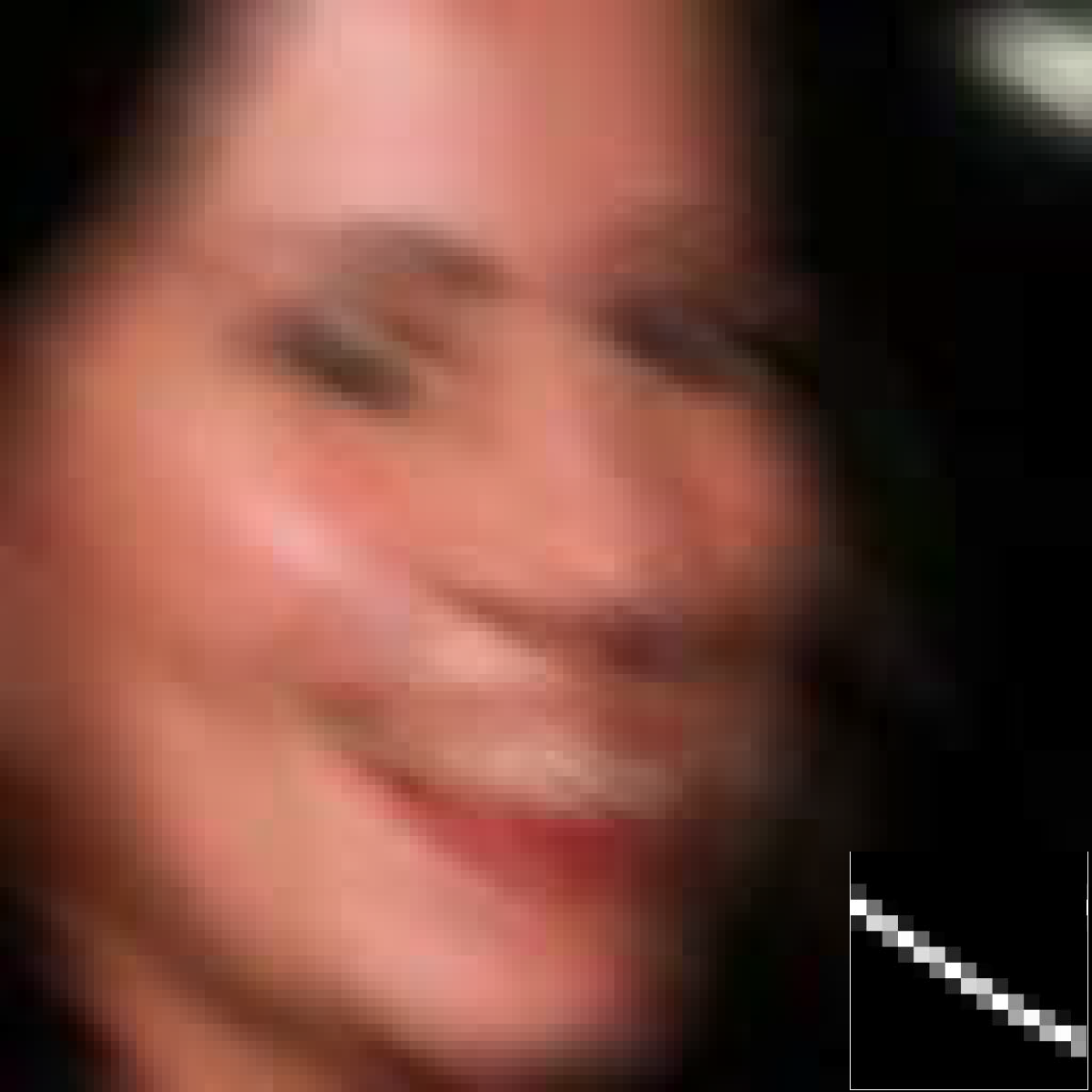} &
            \includegraphics[width=0.09\textwidth]{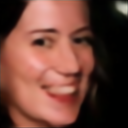} &
			\includegraphics[width=0.09\textwidth]{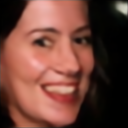} &
			\includegraphics[width=0.09\textwidth]{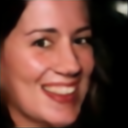} \\
            \includegraphics[width=0.09\textwidth]{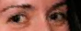} &
			\includegraphics[width=0.09\textwidth]{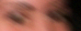} &
			\includegraphics[width=0.09\textwidth]{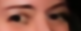} &
            \includegraphics[width=0.09\textwidth]{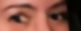} &
			\includegraphics[width=0.09\textwidth]{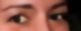}  \\
			(a) & (b) & (c) & (d) & (e)\\
	\end{tabular}
	\vspace{1mm}
	\centering
	\caption{\textbf{Effects of semantic face parsing on image deblurring.}
	  (a) Ground truth images
	  (b) Blurred images
	  (c) Content loss
	  (d) Content loss + global semantic priors
	  (e) Content loss + global semantic priors + local structural losses.
	}
	\label{fig:semantic_deblurring}
	\vspace{-2mm}
\end{figure}

\figref{semantic_deblurring} shows two deblurred results from our Helen test set.
The network optimized solely from the content loss produces overly smooth deblurred results.
The shape of the faces and lips cannot be well recovered as in~\figref{semantic_deblurring}(c).
By introducing the semantic labels as the global priors, the network better reconstructs the outline of faces.
However, the results may not contain fine details in several key components, such as teeth and eyes.
The network with the additional local structural losses restores more details and textures as shown in~\figref{semantic_deblurring}(e).
\tabref{ablation} shows the performance contribution of each component on both the Helen and CelebA test sets.

\Paragraph{Incremental training.}
Real-world blurred images are likely formed by a large diversity of camera motion.
In order to handle random blur kernels in the wild, a simple strategy is to synthesize a large number of blur kernels and blurred images for training.
However, it is difficult to train a deep network from scratch using all blurred images simultaneously as the network has to learn $N$-to-1 mapping where $N$ is the number of blur kernels.
The network may converge to a bad local minimum and cannot restore images well especially for large blur kernels.

To address this issue, we propose a simple yet effective incremental training strategy by incorporating more blur kernels sequentially during training.
We first train the network on smaller blur kernels (i.e., $13 \times 13$).
We then gradually expand the training set by increasing the size of blur kernels.
Specifically, we train the network for $K$ iterations before introducing new blur kernels.
While incorporating new blur kernels, we still sample the existing blur kernels for training until all blur kernels are included.
We set $K = 30000$ iterations in our experiments and train the network for a total of 17 million iterations.

\begin{figure}
	\footnotesize
	\renewcommand{\tabcolsep}{1pt} % adjust horizontal space
	\renewcommand{\arraystretch}{0.5} % adjust vertical space
	\centering
	\begin{tabular}{cccc}
		\includegraphics[width=0.115\textwidth]{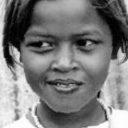} &
		\includegraphics[width=0.115\textwidth]{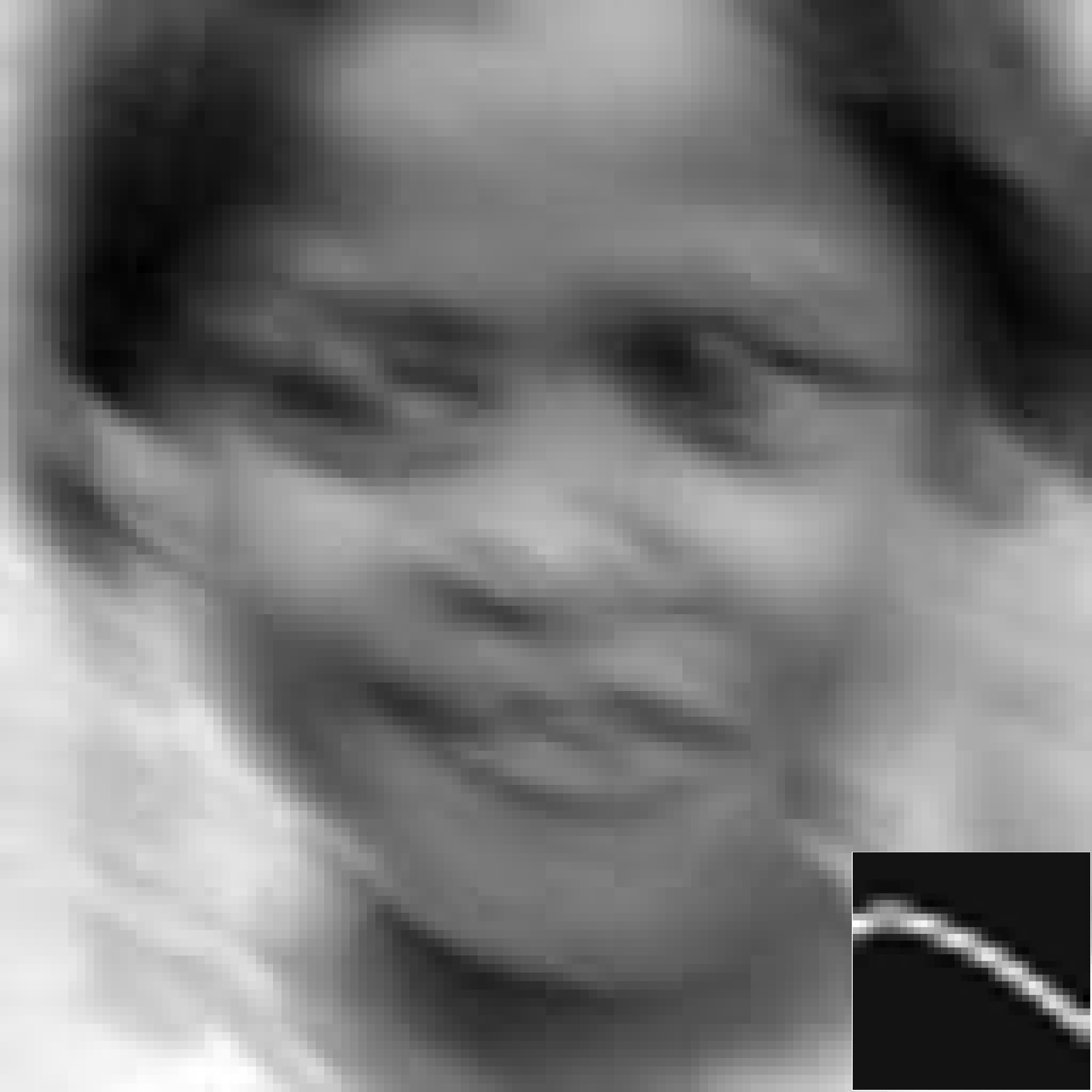} &
		\includegraphics[width=0.115\textwidth]{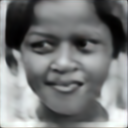} &
		\includegraphics[width=0.115\textwidth]{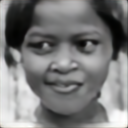}
		\\
		\includegraphics[width=0.115\textwidth]{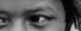} &
		\includegraphics[width=0.115\textwidth]{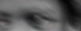} &
		\includegraphics[width=0.115\textwidth]{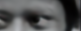} &
		\includegraphics[width=0.115\textwidth]{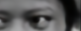}
		\\
		\includegraphics[width=0.115\textwidth]{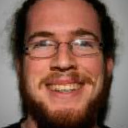} &
		\includegraphics[width=0.115\textwidth]{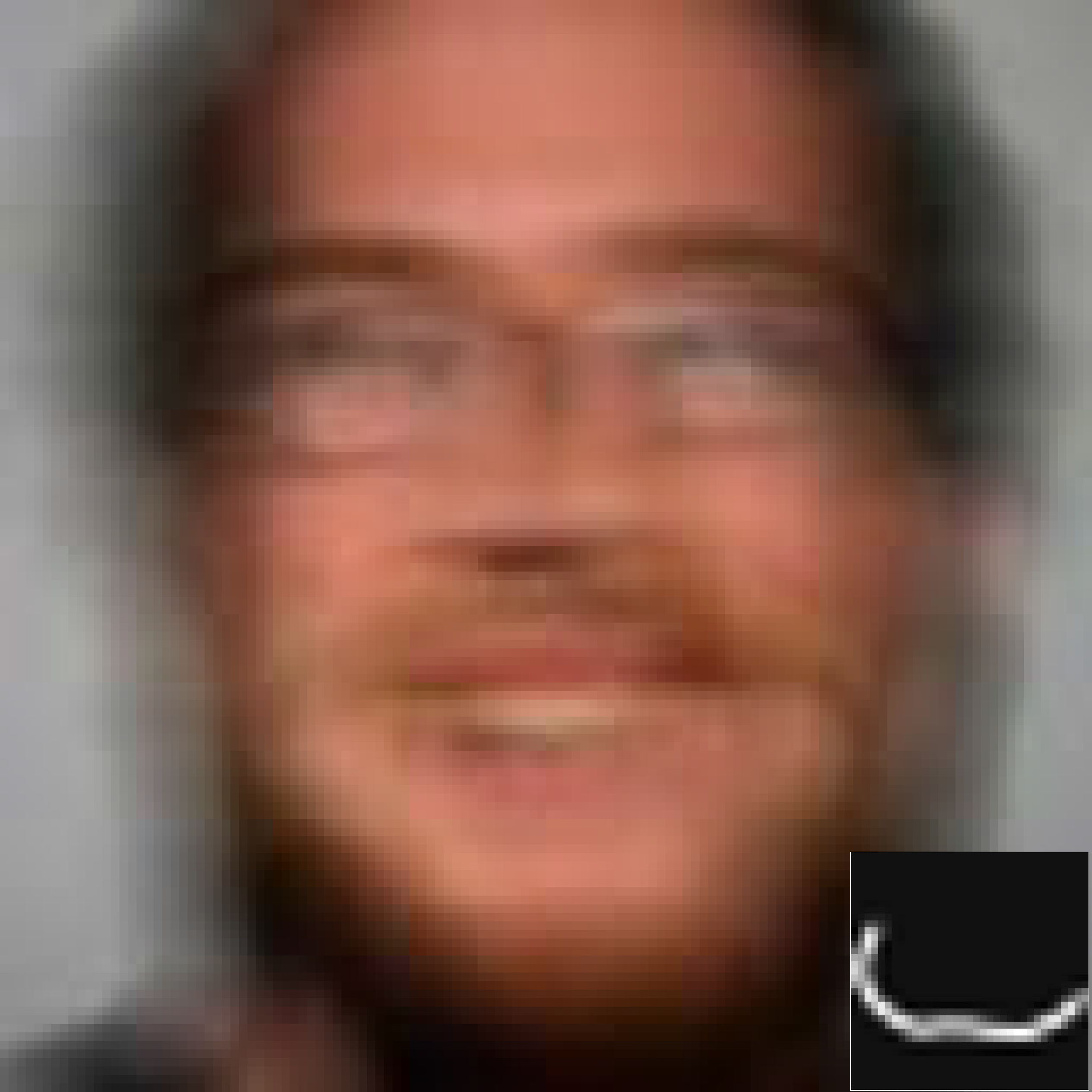} &
		\includegraphics[width=0.115\textwidth]{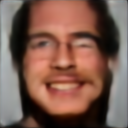} &
		\includegraphics[width=0.115\textwidth]{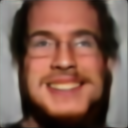}
		\\
		\includegraphics[width=0.115\textwidth]{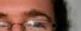} &
		\includegraphics[width=0.115\textwidth]{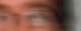} &
		\includegraphics[width=0.115\textwidth]{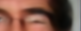} &
		\includegraphics[width=0.115\textwidth]{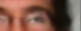}
		\\
		(a) & (b) & (c) & (d)
	\end{tabular}
	%\vspace{1mm}
	\caption{
		\textbf{Visual comparison of training strategies.}
		(a) Ground truth images
		(b) Blurred images
		(c) Direct training
		(d) Incremental training.
	}
	\label{fig:training_strategy_visual}
	\vspace{-4mm}
\end{figure}

\begin{figure}
	\footnotesize
	\renewcommand{\tabcolsep}{1pt}
	\centering
	\begin{tabular}{c}
		\includegraphics[width=0.9\columnwidth]{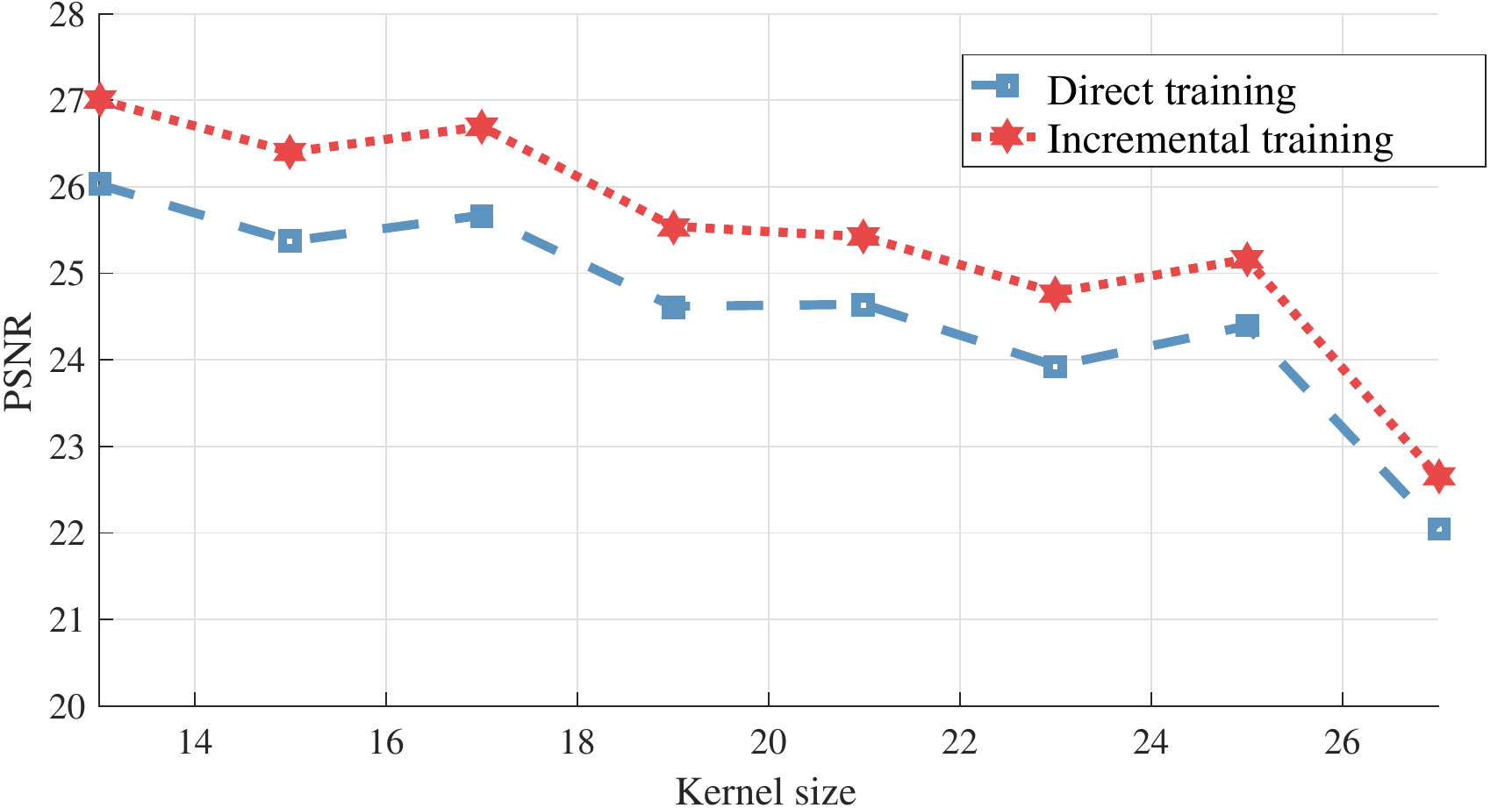}
		%\\
		%\includegraphics[width=0.4\textwidth]{/figure5/5b.pdf}
	\end{tabular}
	\vspace{1mm}
	\caption{
		\textbf{Quantitative evaluation of training strategies.}
		We compare the direct training (blue curve) and the proposed incremental training (red curve) strategies on the Helen test set.
	}
	\label{fig:training_strategy_psnr}
	\vspace{-2mm}
\end{figure}

We provide a comparison of the direct training (i.e., training all blurred kernels simultaneously) and the proposed incremental training in~\figref{training_strategy_visual} and~\tabref{ablation}.
\figref{training_strategy_psnr} shows the quantitative comparison on different sizes of blur kernels.
The proposed incremental training strategy performs better on all sizes of blur kernels and restores the images well.

%\ignorethis{
%}

\begin{table}[t]
	\centering
	\footnotesize
	\caption{
		\textbf{Ablation study}.
		While the model with the perceptual loss achieves the highest PSNR/SSIM, including the adversarial loss produces more realistic face images.
		%We analyze the performance contribution of each component in the proposed model.
		}
	\vspace{1mm}
	\begin{tabular}{c|cc|cc}
		\toprule
		\multirow{2}{*}{Approach} &
		\multicolumn{2}{c|}{Helen} &
		\multicolumn{2}{c}{CelebA} \\
		& PSNR & SSIM & PSNR & SSIM \\
		\midrule
		Content loss
		& 24.85 & 0.849 & 24.23 & 0.864 \\
		+ Global semantic priors
		& 25.32 & 0.857 & 24.32 & 0.864\\
		+ Local structural loss
		& 25.48 & 0.859 & 24.58 & 0.866  \\
        + Incremental training
        & 25.55 & 0.860 & 24.61 & 0.869\\
		+ Perceptual loss
		& \textbf{25.99} & \textbf{0.871} & \textbf{25.05} & \textbf{0.879} \\
		+ Adversarial loss
		& 25.58 & 0.861 & 24.34 & 0.860 \\
		\bottomrule
	\end{tabular}
	\label{tab:ablation}
	\vspace{-2mm}
\end{table}

\begin{table}
	\centering
	\footnotesize
	\caption{
		\textbf{Quantitative comparison with state-of-the-art methods.}
		We compute the average PSNR and SSIM on two test sets.
		}
	\label{tab:performance}
	\vspace{1mm}
	\begin{tabular}{c|cc|cc}
		\toprule
		\multirow{2}{*}{Method} & \multicolumn{2}{c|}{Helen} & \multicolumn{2}{c}{CelebA} \\
		& PSNR & SSIM & PSNR & SSIM \\
		\midrule
		Krishnan et al.~\cite{DBLP:conf/cvpr/KrishnanTF11}
		& 19.30 & 0.670 & 18.38 & 0.672 \\
		Pan et al.~\cite{DBLP:conf/eccv/PanHSY14}
		& 20.93 & 0.727 & 18.59 & 0.677 \\
		Shan et al.~\cite{DBLP:journals/tog/ShanJA08}
		& 19.57 & 0.670 & 18.43 & 0.644 \\
		Xu et al.~\cite{DBLP:conf/cvpr/XuZJ13}
		& 20.11 & 0.711 & 18.93 & 0.685 \\
		Cho and Lee~\cite{DBLP:journals/tog/ChoL09}
		& 16.82 & 0.574 & 13.03 & 0.445 \\
		Zhong et al.~\cite{DBLP:conf/cvpr/ZhongCMPW13}
		& 16.41 & 0.614 & 17.26 & 0.695 \\
		Nah et al.~\cite{Nah_2017_CVPR}
		& 24.12 & 0.823 & 22.43 & 0.832 \\
		Ours
		& \textbf{25.58} & \textbf{0.861} & \textbf{24.34} & \textbf{0.860} \\
		\bottomrule
	\end{tabular}
	\vspace{-3mm}
\end{table}

\begin{figure*}[h!]
	\centering
	\includegraphics[width=1\textwidth]{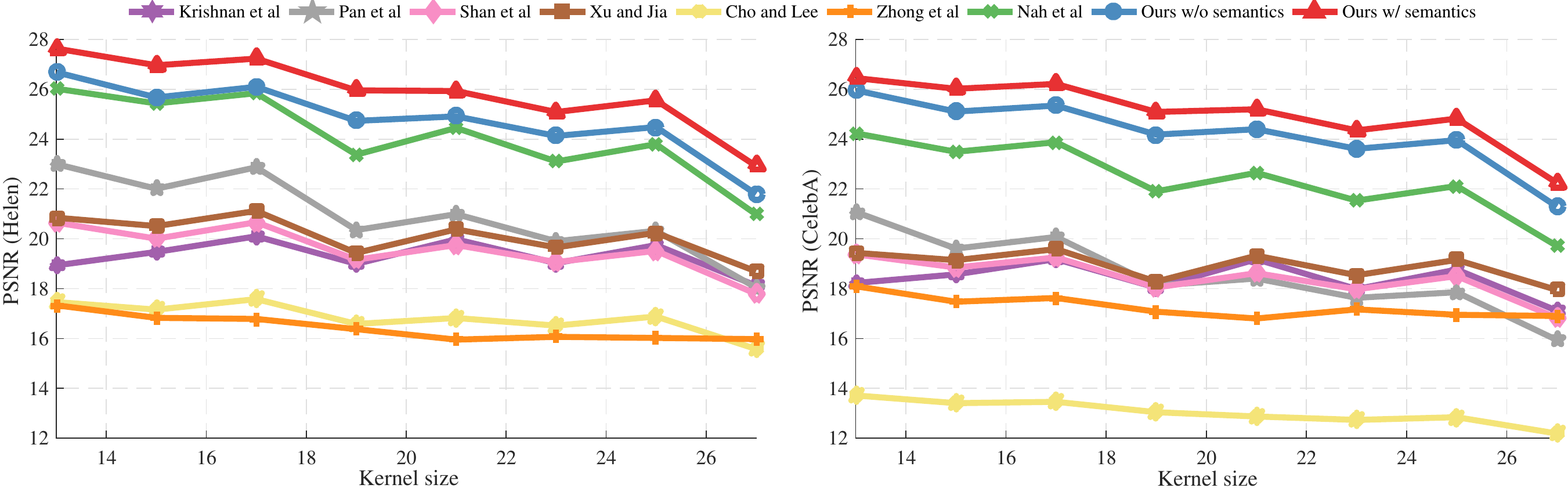}
	\caption{
		\textbf{Quantitative evaluation on different sizes of blur kernels.}
		There are 100 latent clear images, 80 blur kernels and a total of 8,000 blurred images in the Helen and CelebA test sets, respectively.
		Our method performs well on all sizes of blur kernels.
	}
	\label{fig:psnr_ssim_random}
	%\vspace{2mm}
\end{figure*}

\begin{figure*}[h!]
	\footnotesize
	\renewcommand{\tabcolsep}{1pt} % adjust horizontal space
	\renewcommand{\arraystretch}{0.8} % adjust vertical space
	\centering
	\begin{center}
		\begin{tabular}{ccccccccccc}
			\includegraphics[width=0.086\textwidth]{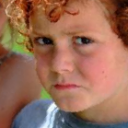}  &
			\includegraphics[width=0.086\textwidth]{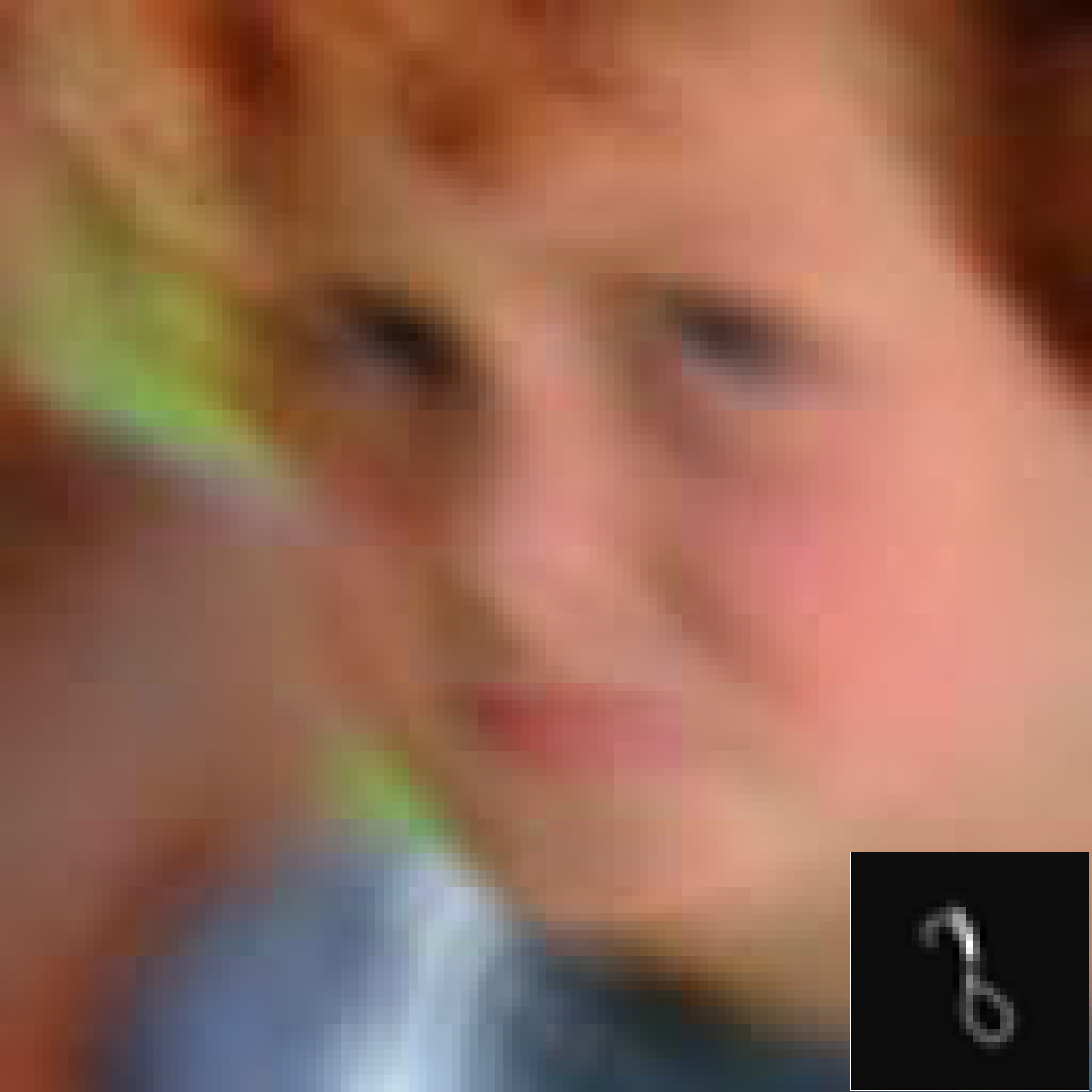} &
			\includegraphics[width=0.086\textwidth]{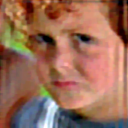} &
			\includegraphics[width=0.086\textwidth]{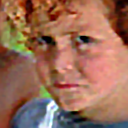} &
			\includegraphics[width=0.086\textwidth]{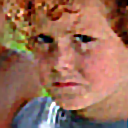} &
			\includegraphics[width=0.086\textwidth]{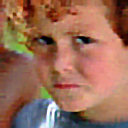} &
			\includegraphics[width=0.086\textwidth]{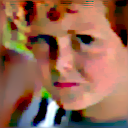} &
			\includegraphics[width=0.086\textwidth]{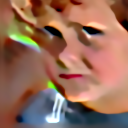} &
			\includegraphics[width=0.086\textwidth]{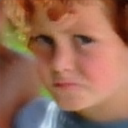} &
			\includegraphics[width=0.086\textwidth]{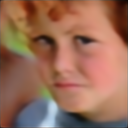} &
			\includegraphics[width=0.086\textwidth]{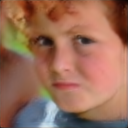} \\
			\includegraphics[width=0.086\textwidth]{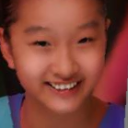} &
			\includegraphics[width=0.086\textwidth]{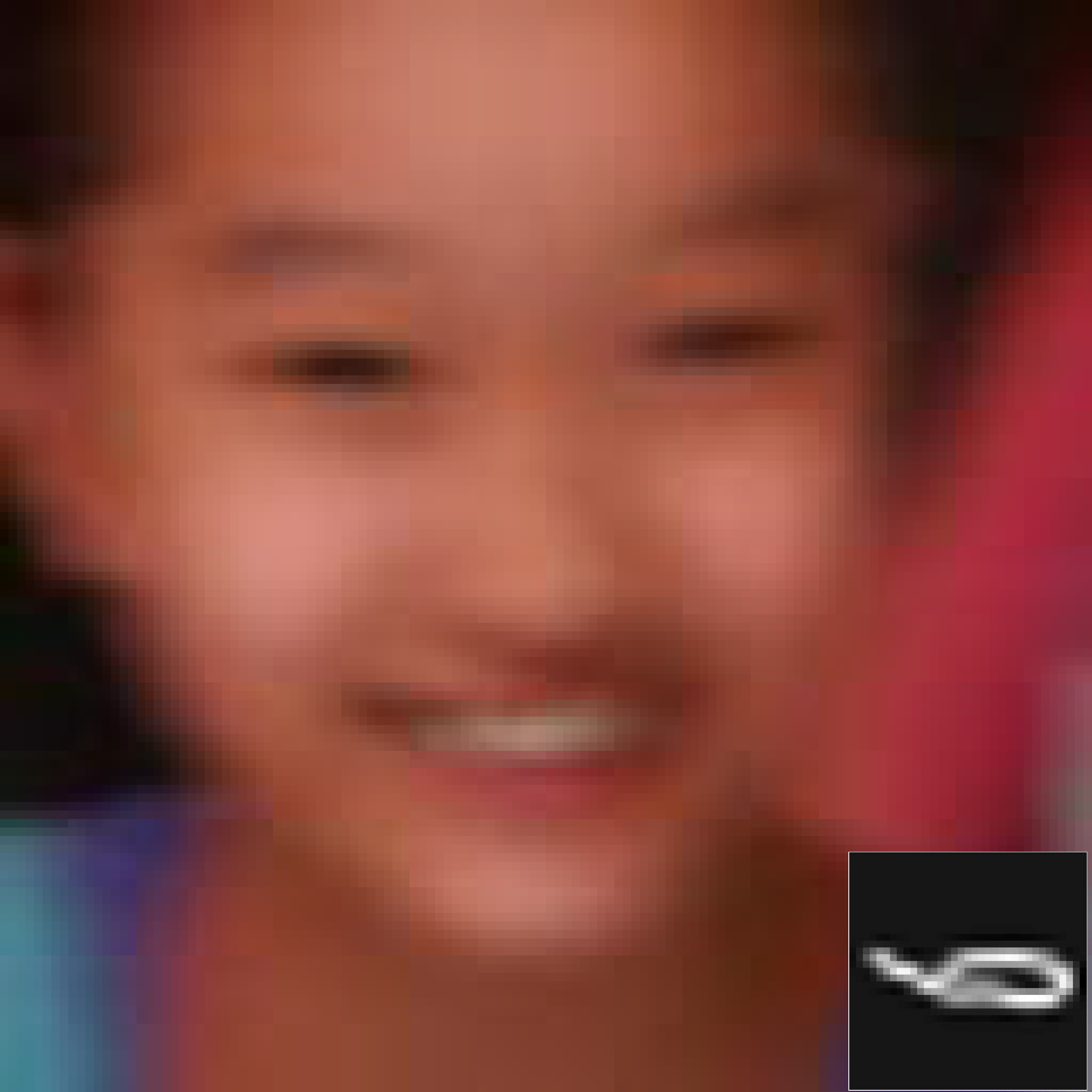} &
			\includegraphics[width=0.086\textwidth]{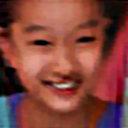} &
			\includegraphics[width=0.086\textwidth]{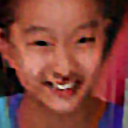} &
			\includegraphics[width=0.086\textwidth]{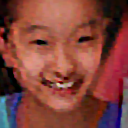} &
			\includegraphics[width=0.086\textwidth]{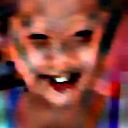} &
			\includegraphics[width=0.086\textwidth]{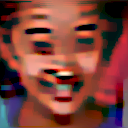} &
			\includegraphics[width=0.086\textwidth]{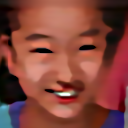} &
			\includegraphics[width=0.086\textwidth]{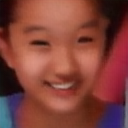} &
			\includegraphics[width=0.086\textwidth]{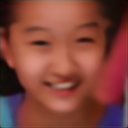} &
			\includegraphics[width=0.086\textwidth]{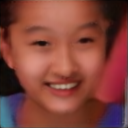} \\
			\includegraphics[width=0.086\textwidth]{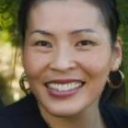} &
			\includegraphics[width=0.086\textwidth]{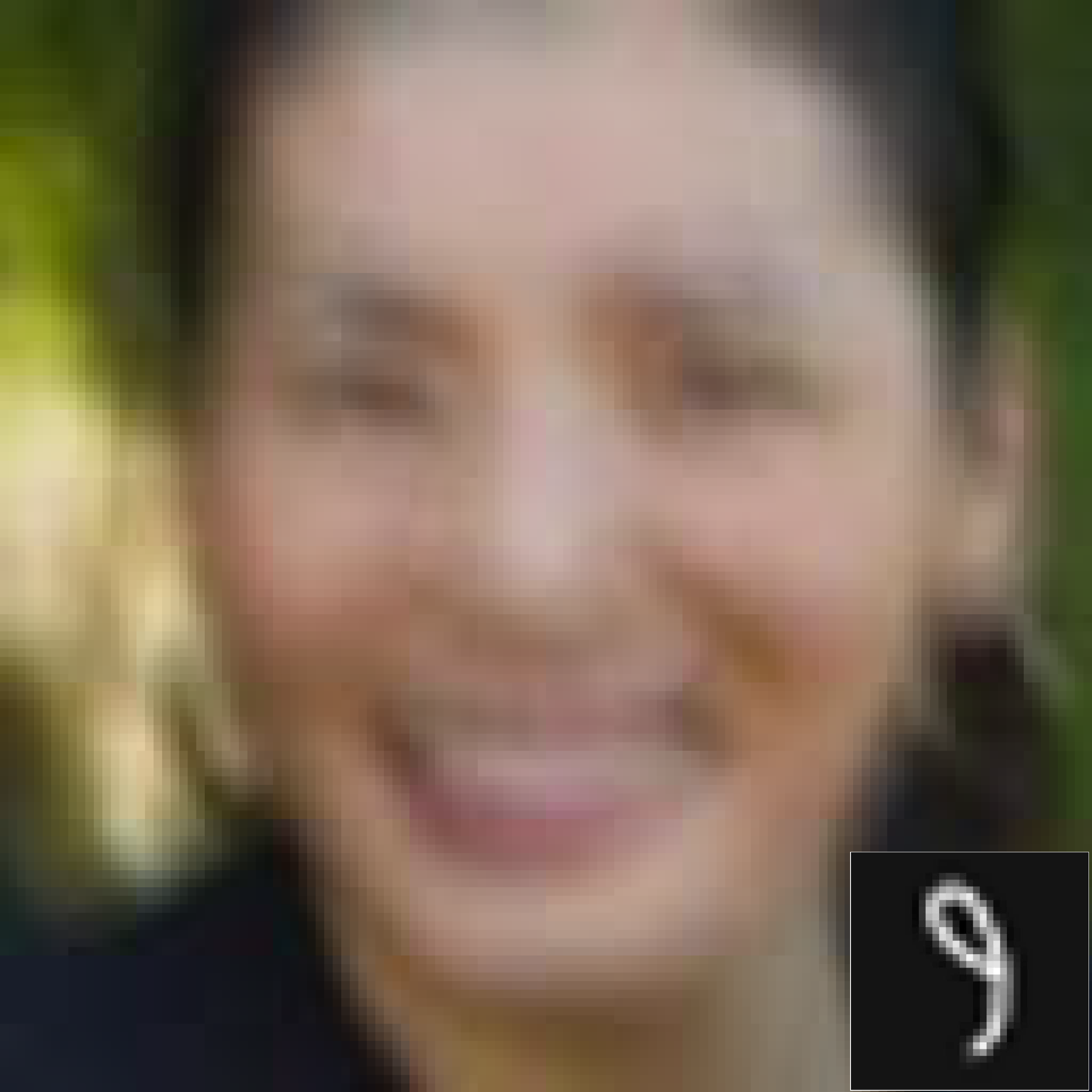} &
			\includegraphics[width=0.086\textwidth]{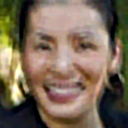} &
			\includegraphics[width=0.086\textwidth]{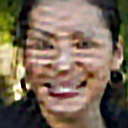} &
			\includegraphics[width=0.086\textwidth]{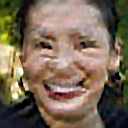} &
			\includegraphics[width=0.086\textwidth]{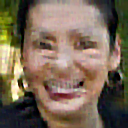} &
			\includegraphics[width=0.086\textwidth]{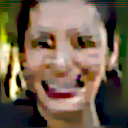} &
			\includegraphics[width=0.086\textwidth]{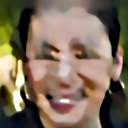} &
			\includegraphics[width=0.086\textwidth]{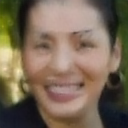} &
			\includegraphics[width=0.086\textwidth]{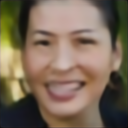} &
			\includegraphics[width=0.086\textwidth]{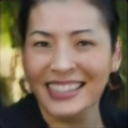} \\
			\includegraphics[width=0.086\textwidth]{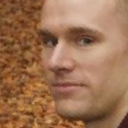} &
			\includegraphics[width=0.086\textwidth]{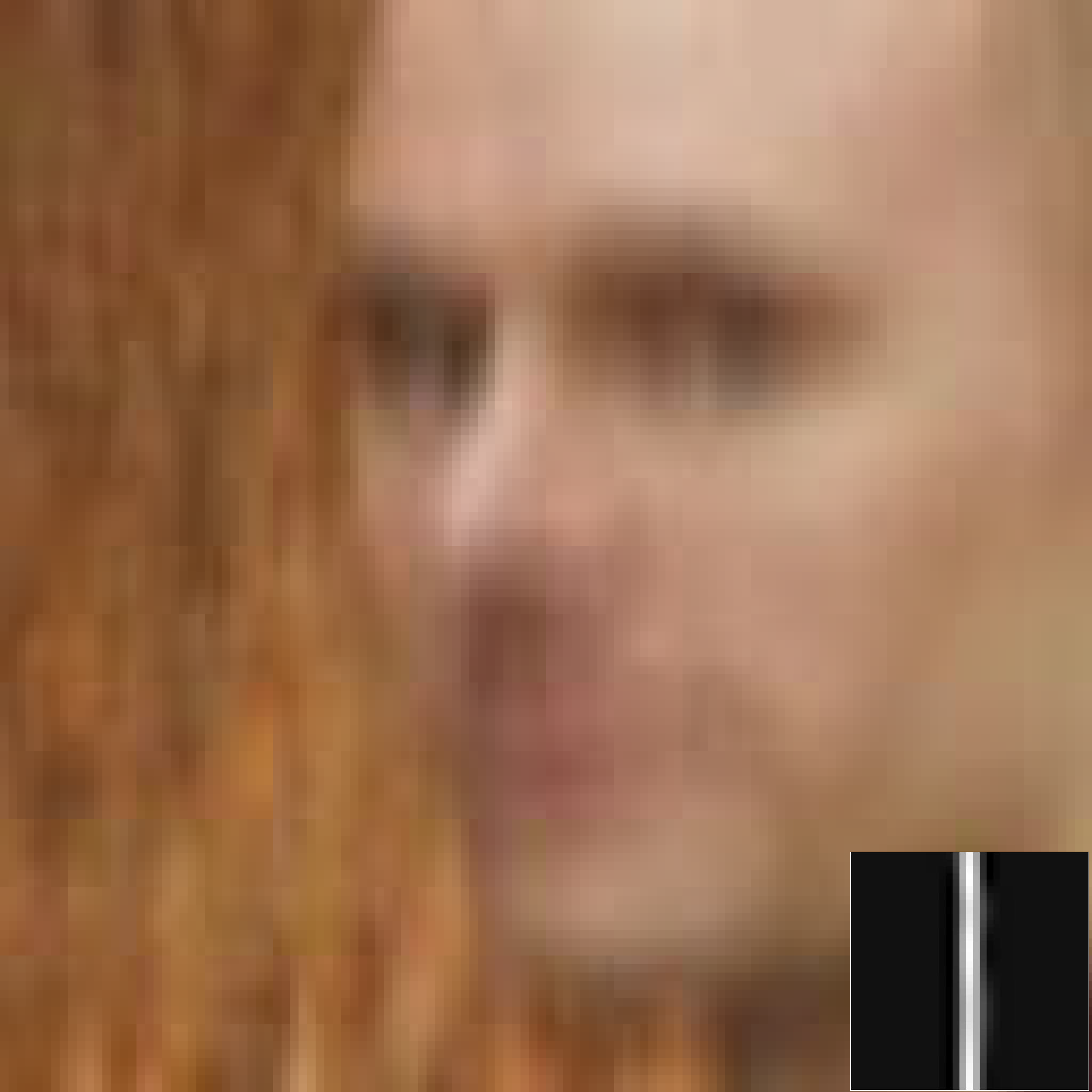} &
			\includegraphics[width=0.086\textwidth]{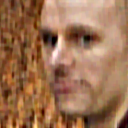} &
			\includegraphics[width=0.086\textwidth]{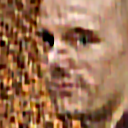} &
			\includegraphics[width=0.086\textwidth]{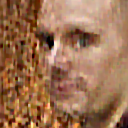} &
			\includegraphics[width=0.086\textwidth]{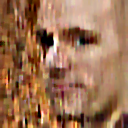} &
			\includegraphics[width=0.086\textwidth]{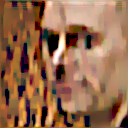} &
			\includegraphics[width=0.086\textwidth]{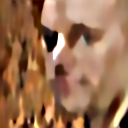} &
			\includegraphics[width=0.086\textwidth]{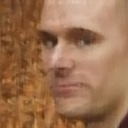} &
			\includegraphics[width=0.086\textwidth]{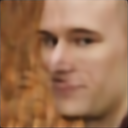} &
			\includegraphics[width=0.086\textwidth]{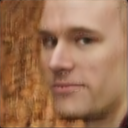}
			\\
			(a) Ground &
			(b) Blurred &
			(c) \cite{DBLP:conf/cvpr/KrishnanTF11} &
			(d) \cite{DBLP:conf/eccv/PanHSY14} &
			(e) \cite{DBLP:journals/tog/ShanJA08} &
			(f) \cite{DBLP:conf/cvpr/XuZJ13} &
			(g) \cite{DBLP:journals/tog/ChoL09} &
			(h) \cite{DBLP:conf/cvpr/ZhongCMPW13} &
			(i) \cite{Nah_2017_CVPR} &
			(j) Ours w/o &
			(k) Ours w/ \\
			truth images & images & & & & & & & & semantics & semantics \\
		\end{tabular}
	\end{center}
	\vspace{-1mm}
	\caption{
		\textbf{Visual comparison with state-of-the-art methods.}
		The results from the proposed method have less visual artifacts and more details on key face components (e.g., eyes and mouths).
	}
	\label{fig:visual_comparison_random}
\end{figure*}

\begin{figure*}
	\centering
	\footnotesize
	\renewcommand{\tabcolsep}{2pt} % adjust horizontal space
	\begin{tabular}{c}
		\includegraphics[width=1.0\textwidth]{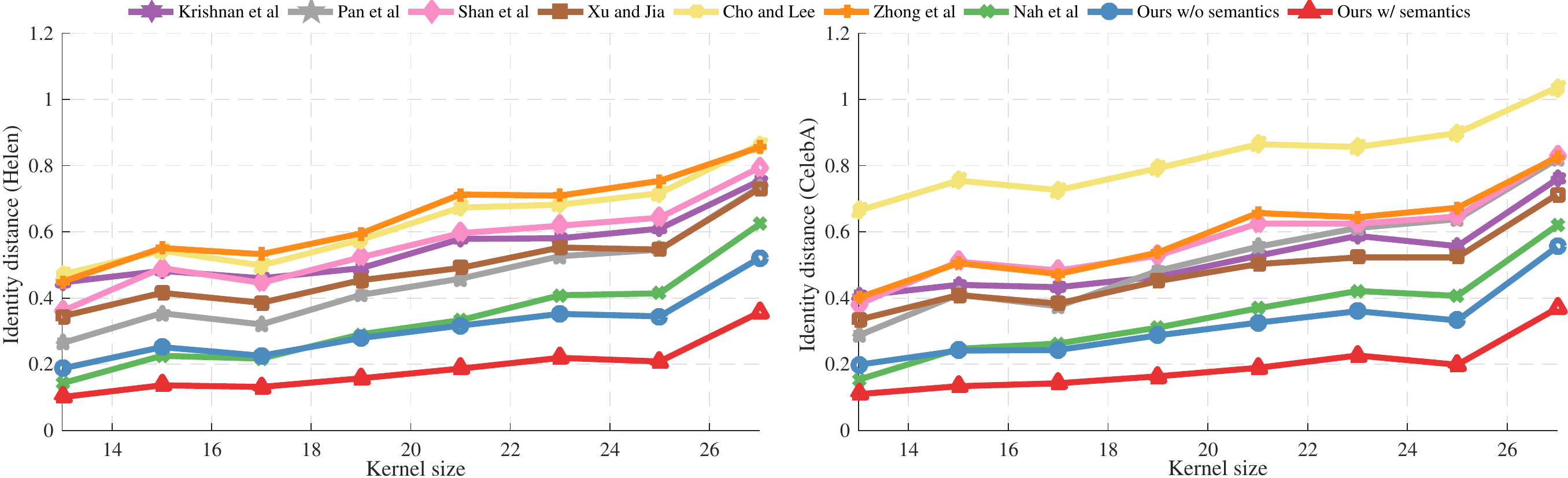}
		\\
	\end{tabular}
	%
	%\vspace{1mm}
	\caption{
		\textbf{Quantitative evaluation on face identity distance.}
		We use the FaceNet~\cite{FaceNet} to compute the identity distance between the clear and deblurred face images.
		The proposed method achieves the lowest identity distance on both the Helen and CelebA test sets.
	}
	\label{fig:identity_distance}
	\vspace{-3mm}
\end{figure*}

\vspace{-3mm}
\Paragraph{Comparisons with state-of-the-arts.}
We provide qualitative and quantitative comparisons with 7 state-of-the-art deblurring algorithms, including MAP-based methods~\cite{DBLP:journals/tog/ChoL09,DBLP:conf/cvpr/KrishnanTF11,DBLP:journals/tog/ShanJA08,DBLP:conf/cvpr/XuZJ13,DBLP:conf/cvpr/ZhongCMPW13}, a face deblurring method~\cite{DBLP:conf/eccv/PanHSY14} and a CNN-based method~\cite{Nah_2017_CVPR}.
We denote our method with all the losses and semantic priors as ``ours w/ semantics'' and our method using only the content loss as ``ours w/o semantics''.

We evaluate the PSNR and SSIM on both the Helen and CelebA datasets in~\tabref{performance}.
\figref{psnr_ssim_random} shows quantitative comparisons on different sizes of blur kernels.
The proposed method performs favorably against the state-of-the-art approaches on both datasets and all blur kernel sizes.
We present visual comparisons in~\figref{visual_comparison_random}.
Conventional MAP-based methods~\cite{DBLP:journals/tog/ChoL09,DBLP:conf/cvpr/KrishnanTF11,DBLP:journals/tog/ShanJA08,DBLP:conf/cvpr/XuZJ13,DBLP:conf/cvpr/ZhongCMPW13} are less effective on deblurring face images and lead to more ringing artifacts.
The MAP-based face deblurring approach~\cite{DBLP:conf/eccv/PanHSY14} is not robust to noise and highly relies on the similarity of the reference image.
The CNN-based method~\cite{Nah_2017_CVPR} does not consider the face semantic information and thus produces overly smooth results.
In contrast, the proposed method utilizes the global and local face semantics to restore face images with more fine details and less visual artifacts.
We provide more visual comparisons in the supplementary material.

\Paragraph{Execution time.}
We analyze the execution time on a machine with a 3.4 GHz Intel i7 CPU (64G RAM) and an NVIDIA Titan X GPU (12G memory).
\tabref{runtime} shows the average execution time based on 10 images with a size of $128 \times 128$.
The proposed method is more efficient than the state-of-the-art deblurring algorithms.

\begin{table}
	\centering
	\footnotesize
	\caption{\textbf{Comparison of execution time.}
		We report the average execution time on 10 images with the size of $128 \times 128$. }
	\label{tab:runtime}
	\begin{tabular}{c|c|c|c}
		\toprule
		Method & Implementation & CPU / GPU & Seconds \\
		\midrule
		Krishnan et al.~\cite{DBLP:conf/cvpr/KrishnanTF11}
		& MATLAB & CPU & 2.52 \\
		Pan et al.~\cite{DBLP:conf/eccv/PanHSY14}
		& MATLAB & CPU & 8.11 \\
		Shan et al.~\cite{DBLP:journals/tog/ShanJA08}
		& C++ & CPU & 16.32 \\
		Xu et al.~\cite{DBLP:conf/cvpr/XuZJ13}
		& C++ & CPU & 0.31 \\
		Cho and Lee~\cite{DBLP:journals/tog/ChoL09}
		& C++ & CPU & 0.41 \\
		Zhong et al.~\cite{DBLP:conf/cvpr/ZhongCMPW13}
		& MATLAB & CPU & 8.07 \\
		Nah et al.~\cite{Nah_2017_CVPR}
		& MATLAB & GPU & 0.09 \\
		Ours
		& MATLAB & GPU & \textbf{0.05} \\
		\bottomrule
	\end{tabular}
	\vspace{-2mm}
\end{table}

\vspace{-2mm}
\Paragraph{Face recognition.}
%
%To further demonstrate the capability of the proposed method on restoring face images, we conduct the experiment of face recognition on deblurred images.
%
We first use the FaceNet~\cite{FaceNet} to compute the identity distance (i.e., the $L_2$ distance on the outputs of FaceNet) between the ground truth face image and deblurred results.
\figref{identity_distance} shows that the deblurred images from the proposed method have the lowest identity distance, which demonstrates that the proposed method preserves the face identity well.

As the CelebA dataset contains identity labels, we conduct another experiment on face detection and identity recognition.
We consider our CelebA test images as a probe set, which has 100 different identities.
For each identity, we collect additional 9 clear face images as a gallery set.
Given an image from the probe set, our goal is to find the most similar face image from the gallery set and identify whether they belong to the same identity.

We use the OpenFace toolbox~\cite{openface} to detect the face for each image in the probe set.
However, due to the motion blur and the ringing artifacts, faces in some of the blurred and deblurred images cannot be well detected.
We then compute the identity distance with all images in the gallery set and select the top-$K$ nearest matches.
We show the success rate of the face detection for blurred images and state-of-the-art deblurring approaches in~\tabref{celebA}.
Furthermore, we compute the recognition accuracy on those successfully detected face images and show the top-1, top-3 and top-5 accuracy.
The proposed method produces fewer artifacts and thus achieves the highest success rate as well as recognition accuracy against other evaluated approaches.

\begin{table}
	\centering
	\footnotesize
	\caption{
		\textbf{Face detection and recognition on the CelebA dataset.}
		We show the success rate of face detection and top-1, top-3 and top-5 accuracy of face recognition.
		%
		%Our method performs favorably against existing deblurring methods on both face detection and recognition.
		%
		%\first{Red} and \second{blue} texts indicate the best and second best performance, respectively.
	}
	\label{tab:celebA}
	\begin{center}
		\begin{tabular}{c|c|c|c|c}
			\toprule
			Method &
			Detection &
			Top-1 &
			Top-3 &
			Top-5
			\\
			\midrule
			Clear images &
			100$\%$ & 71$\%$ & 84$\%$ & 89$\%$
			\\
			\midrule
			Blurred images &
			81$\%$ & 31$\%$ & 46$\%$ & 53$\%$
			\\
			Krishnan et al.~\cite{DBLP:conf/cvpr/KrishnanTF11} &
			84$\%$ & 36$\%$ & 51$\%$ & 59$\%$
			\\
			Pan et al.~\cite{DBLP:conf/eccv/PanHSY14} &
			82$\%$ & 44$\%$ & 57$\%$ & 64$\%$
			\\
			Shan et al.~\cite{DBLP:journals/tog/ShanJA08} &
			80$\%$ & 34$\%$ & 49$\%$ & 56$\%$
			\\
			Xu et al.~\cite{DBLP:conf/cvpr/XuZJ13} &
			86$\%$ & 43$\%$ & 57$\%$ & 64$\%$
			\\
			Cho and Lee~\cite{DBLP:journals/tog/ChoL09} &
			56$\%$ & 21$\%$ & 31$\%$ & 37$\%$
			\\
			Zhong et al.~\cite{DBLP:conf/cvpr/ZhongCMPW13} &
			73$\%$ & 30$\%$ & 44$\%$ & 51$\%$
			\\
			Nah et al.~\cite{Nah_2017_CVPR} &
			90$\%$ & 42$\%$ & 57$\%$ & 64$\%$
			\\
			Ours w/o semantics &
			95$\%$ & 42$\%$ & 55$\%$ & 62$\%$
			\\
			Ours w/ semantics &
			\textbf{99$\%$} & \textbf{54$\%$} & \textbf{68$\%$} & \textbf{74$\%$}
			%7608 & \first{0.5204} & \first{0.6694} & \first{0.7308}
			\\
			\bottomrule
		\end{tabular}
	\end{center}
	\vspace{-4mm}
\end{table}

\begin{figure}
	\footnotesize
	\renewcommand{\tabcolsep}{1pt} % adjust horizontal space
	\renewcommand{\arraystretch}{0.75} % adjust vertical space
	\centering
	\begin{tabular}{ccccc}
		\includegraphics[width=0.09\textwidth]{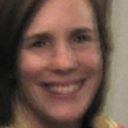} &
		\includegraphics[width=0.09\textwidth]{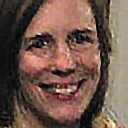} &
		\includegraphics[width=0.09\textwidth]{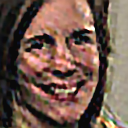} &
		\includegraphics[width=0.09\textwidth]{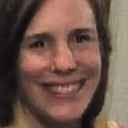} &
		\includegraphics[width=0.09\textwidth]{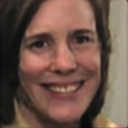}
		\\
		\includegraphics[width=0.09\textwidth]{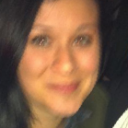} &
		\includegraphics[width=0.09\textwidth]{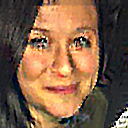} &
		\includegraphics[width=0.09\textwidth]{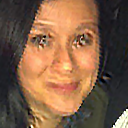} &
		\includegraphics[width=0.09\textwidth]{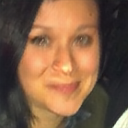} &
		\includegraphics[width=0.09\textwidth]{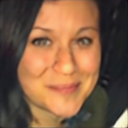}
		\\
		(a) & (b) & (c) & (d) & (e)
	\end{tabular}
	\vspace{1pt}
	\caption{
		\textbf{Visual comparison of real-world blurred images.}
		(a) Blurred images
		(b) Xu et al.~\cite{DBLP:conf/cvpr/XuZJ13}
		(c) Pan et al.~\cite{DBLP:conf/eccv/PanHSY14}
		(d) Nah et al.~\cite{Nah_2017_CVPR}
		(e) Ours
	}
	\label{fig:real_face}
	\vspace{-4mm}
\end{figure}

\vspace{-2mm}
\Paragraph{Real-world blurred images.}
We also test the proposed method on face images collected from the real blurred dataset of Lai et al.~\cite{Lai-CVPR-2016}.
As shown in \figref{real_face}, our method restores more visually pleasing faces than state-of-the-art approaches~\cite{DBLP:conf/eccv/PanHSY14,DBLP:conf/cvpr/XuZJ13}.
We provide more deblurred results of real-world blurred images in the supplementary material.
	
\subsection{Limitations and discussions}
%
%Although the proposed method performs better than existing deblurring algorithms on various sizes of blur kernels, the performance drops significantly when the size of blur kernels becomes larger (e.g., $27 \times 27$).
Our method may fail when the input face image cannot be well aligned, e.g., side faces or extremely large motion blur.
Future work includes improving the performance on handling large and non-uniform blur kernels and relieving the requirement of face alignment.

%extending the proposed method to natural scenes by exploiting scene semantics.

%%%%%%%%%%%%%%%%%%%%%%%%%%%%%%%%%%%%%%%%%%%%%%%%%%%%%%%%%%%%%%%%
%%%%%%%%% Conclusion
%%%%%%%%%%%%%%%%%%%%%%%%%%%%%%%%%%%%%%%%%%%%%%%%%%%%%%%%%%%%%%%%
\section{Conclusions}
\label{sec:conclusions}

In this work, we propose a deep convolutional neural network for face image deblurring.
We exploit the face semantic information as global priors and local structural constraints to better restore the shape and detail of face images.
In addition, we optimize the network with perceptual and adversarial losses to produce photo-realistic results.
We further propose an incremental training strategy for handling random and unknown blur kernels in the wild.
Experimental results on image deblurring, execution time and face recognition demonstrate that the proposed method performs favorably against state-of-the-art deblurring algorithms.

\section*{Acknowledgments}
This work was supported by the Major Science Instrument Program of the National Natural Science Foundation of China under Grant 61527802, and the General Program of National Nature Science Foundation of China under Grants 61371132 and 61471043.

{\small
	\bibliographystyle{ieee}
	\bibliography{deblur}
}
\end{document}